\documentclass[11pt,a4paper]{article}
\usepackage[hyperref]{acl2021}
\usepackage{times}
\usepackage{latexsym}

\usepackage{microtype}

\newcommand*{\affaddr}[1]{#1} % No op here. Customize it for different styles.

\newcommand*{\emailmark}[1][*]{\textsuperscript{#1}}
\newcommand*{\email}[1]{\texttt{#1}}

\aclfinalcopy % Uncomment this line for the final submission
\title{A Transition-based Parser for Unscoped Episodic Logical Forms}

\author{%
Gene Louis Kim\emailmark[$\heartsuit$], Viet Duong\emailmark[$\diamondsuit$], Xin Lu\emailmark[$\spadesuit$], and Lenhart Schubert\emailmark[$\clubsuit$]\\
\affaddr{University of Rochester\\
Department of Computer Science}\\
\email{\{gkim21\emailmark[$\heartsuit$],schubert\emailmark[$\clubsuit$]\}@cs.rochester.edu}\\
\email{\{vduong\emailmark[$\diamondsuit$],xlu32\emailmark[$\spadesuit$]\}@u.rochester.edu}\\
}

\date{}

\usepackage{xspace}
\usepackage{amsmath}
\usepackage{enumitem}
\usepackage{tabularx}
\usepackage{graphicx}
\usepackage{soul}

\newcommand{\codelink}{\url{https://github.com/genelkim/ulf-transition-parser}}
\newcommand{\datasetlink}{\url{https://www.cs.rochester.edu/u/gkim21/ulf/resources/}}
\newcommand{\ulf}[1]{#1} % Formatting for ULFs.
\usepackage{multido}

\newcommand{\extext}[1]{\textit{``#1''}}

\newcommand{\promote}{\textsc{Promote}\xspace}
\newcommand{\promotearc}{\textsc{PromoteArc}\xspace}
\newcommand{\pop}{\textsc{Pop}\xspace}
\newcommand{\gen}{\textsc{Gen}\xspace}
\newcommand{\wordgen}{\textsc{WordGen}\xspace}
\newcommand{\lemmagen}{\textsc{LemmaGen}\xspace}
\newcommand{\tokengen}{\textsc{TokenGen}\xspace}
\newcommand{\namegen}{\textsc{NameGen}\xspace}
\newcommand{\arc}{\textsc{Arc}\xspace}
\newcommand{\push}{\textsc{Push}\xspace}
\newcommand{\sembleu}{\textsc{SemBLEU}\xspace}
\newcommand{\elsmatch}{\textsc{EL-Smatch}\xspace}
\newcommand{\smatch}{\textsc{Smatch}\xspace}

\newcommand{\hb}{\boldsymbol{h}}
\newcommand{\eb}{\boldsymbol{e}}

\begin{document}
\maketitle
\begin{abstract}
``Episodic Logic:~Unscoped Logical Form'' (EL-ULF)
is a semantic representation capturing predicate-argument structure 
as well as more challenging aspects of language % or say "capturing detailed linguistic semantic structure"
within the Episodic Logic formalism. We present the first learned approach
for parsing sentences into ULFs, using a growing set of annotated examples. The results provide a strong baseline for future improvement. Our method learns a sequence-to-sequence model for predicting the transition action sequence within a modified cache transition system. We evaluate the efficacy of type grammar-based constraints, a word-to-symbol lexicon, and transition system state features in this task.
Our system is available at \codelink. We also present the first official annotated ULF dataset at \datasetlink.
\end{abstract}

\section{Introduction}
\vspace{-1mm}
EL-ULF was recently introduced as a semantic representation that accurately captures linguistic %predicate-argument 
semantic
structure within an expressive logical formalism, while staying close to the surface language, facilitating annotation of a dataset that can be used to train a parser~\citep{kim-schubert-2019-type}. The goal is to overcome the limitations of fragile rule-based systems, such as the Episodic Logic~(EL) parser used for gloss axiomatization~\citep{kim-schubert-2016-high} and domain-specific ULF parsers used for schema generation and dialogue systems~\citep{lawley-etal-2019-towards,platonov-etal-2020-spoken}.
EL's rich model-theoretic semantics enables deductive inference, uncertain inference, and natural logic-like inference~\cite{morbini2009LFCR,schubert2000book,schubert2014SP}; and the unscoped version, EL-ULF, supports Natural Logic-like monotonic inferences~\cite{kim-etal-2020-monotonic} and inferences based on some classes of entailments, presuppositions, and implicatures which are common in discourse~\cite{kim-etal-2019-generating}. The lack of robust parsers have prevented large scale experiments using these powerful representations.
%EL-ULF captures linguistic semantic structure in accordance with underlying logical types while leaving word sense, operator scopes, anaphora, and underlying events unspecified.
%Figure~\ref{fig:ulf-ex} shows an example EL-ULF annotation and Section~\ref{sec:ulf} discusses the representational details in more depth.
We will refer to EL-ULF as simply ULF in the rest of this paper.

%ULF is used in schema generation from stories as a rich and formal internal representation of language~\cite{lawley-etal-2019-towards} and has also been shown to support inferences based on some classes of entailments, presuppositions, and implicatures which are common in discourse~\cite{kim-etal-2019-generating}.
%A modern ULF parser would allow the use of these inference-capabilities at a wider scale, replacing currently used fragile and domain-specific rule-based ULF parsers~\cite{kim-schubert-2016-high}.
%ULF is also a critical step in parsing EL formulas.
%EL's rich model-theoretic semantics enables deductive, inference, uncertain inference, and natural logic-like inference~\cite{morbini2009LFCR,schubert2000book,schubert2014SP}. A better ULF parser using modern parsing methods would enable the use of ULF and EL inference-capabilities at a wider scale, overcoming a major source of errors identified by \citet{kim-schubert-2016-high} in a gloss axiomatization system using a domain-specific, rule-based EL parser. 

In this paper we present the first
%published
system that learns to parse ULFs of English sentences from an annotated dataset, and provide the first official release of the annotated ULF corpus, whereon our system is trained. We evaluate the parser using \sembleu~\citep{song-gildea-2019-sembleu} and a modified version of \smatch~\citep{cai-knight-2013-smatch}, establishing a baseline for future work.

An initial effort in learning a parser producing a representation as rich as ULF
is bound to face a data sparsity issue.% 
% Data sparsity is a major issue in learning a parser from this first ULF corpus release.
\footnote{The training set in our initial release is only 1,378 sentences.}
%In comparison, SemEval 2016 task 8 on AMR parsing~\citep{may-2016-semeval} was considered relatively data-sparse with 16,833 sentences in the training set.
Thus a major goal in our choice of a transition-system-based parser has been to reduce the search space of the model. We investigate three additional methods of tackling this issue: (1)~constraining actions in the decoding phase based on faithfulness to the ULF type system, (2)~using a lexicon to limit the possible word-aligned symbols that the parser can generate, and (3)~defining learnable features of the transition system state.
%which are supplied to the decoder at each step.
%\vspace{-1mm}
\begin{figure}[t]
    \centering
    {\small
    \begin{minipage}{0.95\linewidth}
        %\ulf{(|Ali| ((pres do.aux-s) not (know.v (that\\
        %\hspace*{1em}(i.pro (work.v (adv-a (with.p
        %(a.d dog.n)))))))))}
        \ulf{(i.pro ((pres want.v)\\
        \hspace*{2.7em}(to (dance.v\\
        \hspace*{4.1em}(adv-a (in.p (my.d ((mod-n new.a)\\
        \hspace*{11.9em}(plur shoe.n)))))))))}
    \end{minipage}
    }
    \vspace{-2mm}
    \caption{An example ULF for the sentence, \extext{I want to dance in my new shoes}.}
    \label{fig:ulf-ex}
\vspace{-5mm}
\end{figure}
\vspace{-1mm}
\section{Unscoped Logical Form}
\label{sec:ulf}
\vspace{-1mm}
Episodic Logic is an extension of first-order logic~(FOL) that closely matches the form and expressivity of natural language, using reifying operators to enrich the domain of basic individuals and situations with propositions and kinds, keeping the logic first-order. It also uses other type-shifters, e.g., for mapping predicates to modifiers, and allows for generalized quantifiers~\citep{schubert2000book2}. 
ULF fully specifies the semantic type structure of EL by marking the types of the atoms and all of the predicate-argument relationships while leaving operator scope, anaphora, and word sense unresolved~\citep{kim-schubert-2019-type}. ULF is the critical first step to parsing full-fledged EL formulas. Types are marked on ULF atoms with a suffixed tag resembling the part-of-speech (e.g., \ulf{.v}, \ulf{.n}, \ulf{.pro}, \ulf{.d} for verbs, nouns, pronouns, and determiners, respectively). Names are instead marked with pipes (e.g. \ulf{\textbar John\textbar}) and a closed set of logical and macro operators have unique types and are left without a type marking. Each suffix denotes a set of possible semantic denotations, e.g. \ulf{.pro} always denotes an \textit{entity} and \ulf{.v} denotes an \textit{n-ary predicate} where \textit{n} can vary. The symbol without the suffix or pipes is called the \textit{stem}.

Type shifters in ULF maintain coherence of the semantic type compositions. For example, the type shifter \ulf{adv-a} maps a predicate into a verbal predicate modifier as in the prepositional phrase \extext{in my new shoes} in Figure~\ref{fig:ulf-ex}, as opposed to its predicative use \extext{A spider is in my new shoes}.
%\todo{Maybe highlight the type shifters in the figure? Maybe include a tree diagram version of a subtree?}%% if space permits...
%-GK-% Doesn't look like we'll have any extra space.

The syntactic structure is closely reflected in ULF even under syntactic movement through the use of rewriting \textit{macros} which explicitly mark these occurrences and upon expansion make the exact semantic argument structure available. Also, further resembling syntactic structure, ULFs are trees. The operators in operator-argument relations of ULF can be in first or second position, disambiguated by the types of the participating expressions. This further reduces the amount of word reordering between English and ULFs. The EL type system only allows function application for combining types, $\langle A, B\rangle, A \rightarrow B$, much like Montagovian semantics~\citep{montague1970}, but without type-raising.

\vspace{-1mm}
\section{Background}
\vspace{-1mm}
%Currently, there is semantic parsing research occurring on multiple representational fronts, which is showcased by the cross-framework meaning representation parsing task~\cite{oepen-etal-2019-mrp}. This task includes AMRs~\cite{banarescu2013abstract}, Discourse Representation Structures~\cite{kamp1981FMSL} in the Parallel Meaning Bank~\cite{abzianidze-etal-2017-parallel}, Universal Conceptual Cognitive Annotation graphs~\cite{abend-rappoport-2013-universal}, Prague Tectogrammatical Graphs~\cite{bejcek-etal-2012-prague}, and Elementary Dependency Structures~\cite{oepen-lonning-2006-discriminant}. ULF is unique in among these as being directly tied to a model-theoretic logic with expressive capacity beyond FOL.

Currently, there is semantic parsing research occurring on multiple representational fronts, which is showcased by the cross-framework meaning representation parsing task \cite{oepen-etal-2019-mrp}.
%The 2020 edition of this task includes AMRs \cite{banarescu2013abstract}, Discourse Representation Structures (DRS) \cite{kamp1981FMSL} in the Parallel Meaning Bank \cite{abzianidze-etal-2017-parallel}, Universal Conceptual Cognitive Annotation graphs~\cite{abend-rappoport-2013-universal}, Prague Tectogrammatical Graphs \cite{bejcek-etal-2012-prague}, and Elementary Dependency Structures (EDS) \cite{oepen-lonning-2006-discriminant}.\footnote{We cite the 2019 edition of this task since citation information for the 2020 edition is not yet available.}
The key differentiating factor of ULF from other meaning representations is the model-theoretic expressive capacity. To highlight this, here are a few limitations of notable representations: AMR neglects issues such as articles, tense, and nonintersective modification in favor of a canonicalized form that abstracts away from the surface structure; Minimal Recursion Semantics \citep{copestake2005RLC} captures meta-level semantics for which inference systems cannot be built directly based on model-theoretic notions of truth and entailment; and DRSs have the same expressive power as FOL which precludes phenoma such as generalized quantifiers, modification, and reification. Due to space limitations, we refer to \citet{kim-schubert-2019-type} for an in-depth description and motivation of ULF, including comparisons to other representations. We also refer to \citet{schubert2015AAAI} which places EL---the antecedent of ULF---in a broad context.

Our ULF parser development draws inspiration from the body of semantic parsing research on graph-based formalism of natural language, in particular, the recent advances in AMR parsing \cite{peng-etal-2018-sequence,zhang-etal-2019-amr}. The core organization of our parser is based on \citet{peng-etal-2018-sequence}, which uses a sequence-to-sequence model to predict the transition action sequence for a cache transition system with transition system features and hard attention alignment.

There are many transition-based parsers that were developed for parsing meaning representations~\cite{zhang-etal-2016-transition-based,buys-blunsom-2017-robust,damonte-etal-2017-incremental,hershcovich-etal-2017-transition}. These are mainly based on what's called an arc-eager parsing method, termed by \citet{abney1991memory}. Arc-eager parsing greedily adds edges between nodes before full constituents are formed, which keeps the partial graph as connected as possible during the parsing process \citep{nivre2004incrementality}. They modify arc-eager parsing in various ways to generalize to the graph structures. Our transition system can be considered a modification of bottom-up arc-standard parsing due to restrictions on arc formation. While this leads to a longer action sequence for parsing, the parser's access to complete constituents allows promotion-based symbol generation for unary operators such as type shifters and standard bottom-up type analysis for constrained parsing.

\vspace{-1mm}
\section{Our Transition System}
\vspace{-1mm}
\label{sec:transition-system}
\begin{figure}
%\vspace{-15pt}
%\includegraphics[width=\linewidth]{img/transition-system-diagram.png}
%\vspace{-15pt}
\centering
\includegraphics[width=0.9\linewidth]{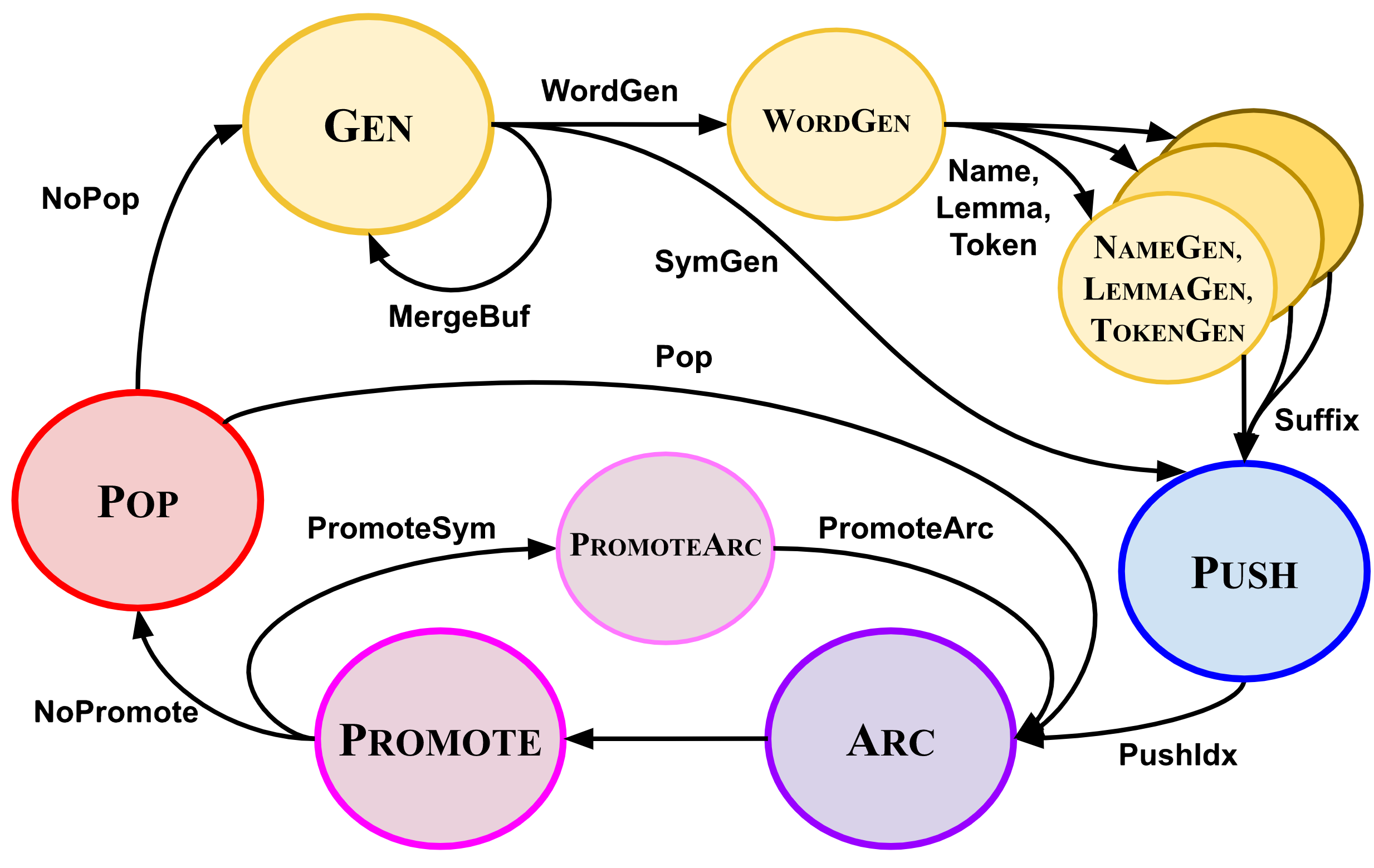}
\vspace{-3mm}
\captionof{figure}{State transition diagram of the node generative transition system. Nodes in the figure are phases and edges are actions. %*Gen is any \textit{WordGen} action or the \textit{SymGen} action.
An unlabeled edge means that this state transition occurs no matter what action is taken in that phase. The transition system starts in the \gen phase.}
\vspace{-6mm}
\label{fig:transition-system}
\end{figure}

Our transition system is a modification of the cache transition system~\cite{gildea-etal-2018-cache} which has been shown to be effective in AMR parsing~\cite{peng-etal-2018-sequence}. The distinctive aspect of our version is that the transition system generates nodes that are derived, but distinct, from the input sequence. We call it a node generative transition system. This eliminates the two-stage parsing framework of \citet{peng-etal-2018-sequence}. Our transition system also restricts the parses to be bottom-up to enable node generation and decoding constraints by the available constituents since ULF has an bottom-up compositional type system. The transition parser configuration is
\vspace{-4mm}
\begin{equation}
    C = (\sigma, \eta, \beta, G_p)
\vspace{-3mm}
\end{equation}
where $\sigma$ is the stack, $\eta$ is the cache, $\beta$ is the buffer, and $G_p$ is the partial graph. The parser is initialized with ($[]$,$[\$,\ldots,\$]$,$[w_1,\ldots,w_n]$,$\emptyset$), that is an empty stack, the cache with null values~($\$$), the buffer with the input sequence of words, where each word is a token, lemma, POS tuple, $w_i = (t_i, l_i, p_i)$, and an empty partial graph, $G_p = (V_p, E_p)$, where $V_p$ is ordered. A vertex, $v_i = (s_i, a_i) \in V$, is a pair of a ULF symbol $s_i$, and its alignment $a_i$---the index of the word from which $s_i$ was produced. We will refer to the leftmost element in $\beta$ as $w_\text{next}$.

While the size of the cache is a hyperparameter that can be set for the cache transition parser, we restrict the cache size to 2 in order to keep the oracle simple despite the newly added actions. This means that only tree structures can be parsed. In describing the transition system, we differentiate between \textit{phases} and \textit{actions}. Phases are classes of states in the transition system and the actions move between states. Figure~\ref{fig:transition-system} shows the full state transition diagram and shows how the phases dictate which actions can be taken and how actions move between phases. Actions may take variables to specify how to move into the next phase. Phases also determine which features go into the determining the next action. We will write phases in small caps (e.g. \gen) and actions in bold (e.g. \textbf{TokenGen}) for clarity.

The \gen and \promote phases are novel to our transition system. The \gen phase generates graph vertices that are transformations of the buffer values. This allows us to put words of the input sentence in $\beta$ instead of a pre-computed ULF atom sequence. The \promote phase enables context-sensitive symbol generation. It generates unaligned symbols in the context of an existing constituent in the partial graph. (Use of logical operators without word alignments only makes sense with respect to something for the operators to act on.)
%In our modified cache transition system, the buffer elements are words, $\eta = [w_1,...,w_n]$, where a word is a token, lemma, POS tuple, $w_i = (t_i, l_i, p_i)$. 
%The vertex set, $V$ from $G = (V, E)$, is ordered and the a vertex, $v_i = (s_i, a_i) \in V$, is a pair of a ULF symbol $s_i$, and its alignment $a_i$---the index of the word from which $s_i$ was produced. 
We now describe each of the actions in the transition system. 
The following are parser actions that were almost directly inherited from the vanilla cache transition parser.
\vspace{-3mm}
\begin{itemize}[leftmargin=*, noitemsep]
    \item \textbf{PushIndex($i$)} pushes $(i,v)$ onto $\sigma$, where $v$ is the vertex currently at index $i$ of $\eta$. Then it moves the vertex generated by the prior \gen phase to index $i$ in $\eta$. 
    
    \item \textbf{Arc($i,d,l$)} forms an arc with label $l$ in direction $d$~(i.e. left or right) between the vertex at index $i$ of the cache and the rightmost vertex in the cache.

    \item \textbf{Pop} pops $(i,v)$ from $\sigma$ where $i$ is the index of $\eta$ which $v$ came from. $v$ is placed at index $i$ of $\eta$ and shifts the appropriate elements to the right.
\end{itemize}
\vspace{-3mm}

\noindent
We introduce two sets, $S_p$ and $S_s$, which define the vocabulary of the two unaligned symbol generation actions: \textbf{PromoteSym} and \textbf{SymGen}, respectively. $S_p$ consists of logical and macro operators that do not align with English words. $S_s$ consists of symbols that could not be aligned in the training set and are not members of $S_p$.
\vspace{-1mm}
\subsection{Promotion-based Symbol Generation}
\label{sec:promote-phase}
\vspace{-1mm}
%% I felt the following needed to be said above, where Promote is introduced...-Len
%% We add the \promote phase to the parser in order to generate unaligned symbols in the context of an existing constituent in the partial graph. The motivation is that use of logical operators without word alignments only make sense with respect to something to act on.
%
%This aims to model this coupling in the transition system.
%The parser first decides, which symbol $s \in S_p$, if any, to generate. Then if a symbol was generated the parser chooses the arc label from $s$ to $s_{c_r}$.
%whether or not the current rightmost symbol $s$ in the cache has a reserved symbol as its parent and second, what the arc label will be between the reserved symbol and current symbol $s$.
\promote includes a subordinate \promotearc phase for modularizing the parsing decision. The following parsing actions are in this phase.
\vspace{-3mm}
\begin{itemize}[leftmargin=*, noitemsep]
    \item \textbf{PromoteSym($s_p$)} generates a promotion symbol, $s_p \in S_p$, appends the vertex $(s_p, \text{NONE})$ to $V_p$, and proceeds to the \promotearc phase.
    %it to the end of both the symbol list and the reserved symbol list of the graph constructed so far. A reserved symbol is a symbol in the reserved set specified when initializing the parser. The generated symbol $s$ is then ready to be processed further.
    \item \textbf{NoPromote} skips the \promote phase and proceeds to the \pop phase.
    %-- This action notifies the parser that at this point, no promote action should be made on current symbol and therefore the promote phase is skipped.
    \item \textbf{PromoteArc($l$)} makes an arc from the last added vertex, $v_p$, to the vertex at the rightmost position of the cache, $v_{\eta_r}$, by adding $(v_p, v_{\eta_r}, l)$ to $E_p$. $v_p$ then takes the place of $v_{\eta_r}$ in the cache and $v_{\eta_r}$ is no longer accessible by the transition system. The system proceeds to the \arc phase.
    %with and arc from $v$ to $c_r$ with label $l$. The vertex corresponding to the promoted symbol then replaces that corresponding the rightmost symbol in the cache for further processing. This action is special for Arc action between reserved symbols and other symbols.
\end{itemize}

\vspace{-1mm}
\subsection{Sequential Symbol Generation}
\label{sec:gen-phase}
\vspace{-1mm}
%In order to encapulate the full parsing process in the transition system, we replace the \textbf{Shift} action in
We replace the \textbf{Shift} action with the \gen phase to generate ULF atoms based on the tokenized text input. This phase allows the parser to generate a symbol using $w_\text{next}$ as a foundation, or generate an arbitrary symbol that is not aligned to any word in $\beta$. \gen includes subordinate phases \wordgen, \lemmagen, \tokengen, and \namegen for modularizing the decision process.
%The parser will decide to generate a reserved symbol, an aligned(i.e. non-reserved) symbol or merge buffer. Reserved symbol is generated through seqGen or symGen, with aligned symbol is generated through conGen. Otherwise, no symbol is generated, but the first two words in the buffer will be merged through mergeBuf.
\vspace{-3mm}
\begin{itemize}[leftmargin=*, noitemsep]
    % \setlength\itemsep{-1mm}
    %\item \textbf{WordGen($s$)} generates a ULF symbol using the input text as the stem and is decomposed into three subtypes.
    \item \textbf{WordGen} proceeds to \wordgen phase, in which the following actions are available.
    % \vspace{-1mm}
    \begin{enumerate}[leftmargin=*,noitemsep,topsep=0pt]
    \item \textbf{Name} proceeds to the \namegen phase. 
    %stores the token as a ULF name symbola stem (e.g. |John|).
    \item \textbf{Lemma} proceeds to the \lemmagen phase. 
    %stores the lemma as a ULF atom stem.
    \item \textbf{Token} proceeds to the \tokengen phase.
    %store the token as a ULF atom stem.
    \end{enumerate}
    \item \textbf{Suffix($e$)} is the only action available in the \namegen, \lemmagen, and \tokengen phases. 
    %% I can't figure out the following sentence. I can perhaps read it as follows, but don't know if that makes sense: "It generates a ULF symbol $s$ consisting of a stem and suffix extension $e$; in the \namegen phase, the stem is the text token; in the \tokengen phase, the basic stem is the token; and in the \lemmagen phase, the basic stem is the lemma derived from the token." BUt I don't know how "stem" and "basic stem" differ. -Len
    %PLEASE REWORD THE FOLLOWING\\
    It generates a symbol $s$ consisting of a stem and suffix extension $e$ from %the leftmost word in $\beta$,
    $w_\text{next}$. In the \namegen phase, the stem is $t_\text{next}$ with surrounding pipes; in the \tokengen phase, the stem is $t_\text{next}$; and in the \lemmagen phase, the stem is $l_\text{next}$.
    %with the token as the ULF stem of a name if in the \namegen phase, the token as a basic ULF stem if in the \tokengen phase, and the lemma as the basic stem if in the \lemmagen phase.
    $(s, i)$ where $i$ is the index of $w_\text{next}$ 
    %the leftmost word of $\beta$~(the aligned word) 
    is added to $V_p$ and we move forward one word in $\beta$. The system proceeds to the \push phase.
    \item \textbf{SymGen($s$)} adds an unaligned symbol $(s, \text{NONE})$ to $V_p$ and proceeds to the \push phase.
    \item \textbf{SkipWord} skips word in $\beta$ and returns to the \gen phase.
    %without an aligned word, directly into the symbol sequence.
    %This generates unaligned symbols that the promote action cannot generate, say because the symbol has not descendants. 
    %\item \textbf{SeqGen($s$)} -- This action generates a reserved symbol belongs to the list of in-sequence symbols specified, if any. If the symbol has its entire constituent being in-sequence reserved symbols or promote reserved symbol, then the uppermost in-sequence symbol will be generated.
    \item \textbf{MergeBuf} takes $w_\text{next}$ and merges it with the word after it $w_{\text{next}+1}$.
    %the current token at the front of the buffer, $v_\beta$, and merges it with the token after it $v_{\beta+1}$.
    This is stored at the front of the buffer as a pair $(v_\beta, v_{\beta+1})$. This forms a single stem with a space delimiter in the \namegen phase and an underscore delimiter in the \lemmagen and \tokengen phases. The system returns to the \gen phase. This is used to handle multi-word expressions (e.g. \extext{had better}).
    %merges two words in the buffer into one. Any further processing will treat the merged words as one. This action is usually followed by an attempt to generate symbol related to the combined words.
\end{itemize}
\vspace{-3mm}
%Figure~\ref{fig:transition-system} is a state transition diagram showing how the actions connect the different phases together. 
The transition system begins in the \gen phase.
% TODO(gene)
% Perhaps a concrete example is better here ... ? I agree, but I think that will take up more space. 
%Given a sentence, the transition parser will look at the leftmost word in the buffer, either generating a corresponding symbol or merging the first and second leftmost words in the buffer. If two words are merged, the parser will return to the initial state and retry generating a symbol for the combined word. If a symbol is generated, the parser will shift it out of the buffer and attempt to make new arcs between this symbol and other symbols in the cache. Then the parser will proceed into the \promote phase if the current symbol has a reserved symbol as its parent. Once the phase is completed, another decision on making new arcs will be made. Otherwise, the parser will proceed to the next word in the buffer.

\vspace{-1mm}
\subsection{Oracle Extraction Algorithm}
\vspace{-1mm}
In order to train a model of the parser actions, we need to extract the desired action sequences from gold graphs. We modify the oracle extraction algorithm for the vanilla cache transition parser, described by \citet{gildea-etal-2018-cache}. The oracle starts with a gold graph $G_g = (V_g, E_g)$ and maintains the partial graph $G_p = (V_p, E_p)$ of the parsing process, where $V_g$ is sequenced by the preorder traversal of $G_g$. The oracle maintains $s_{\text{next}}$, the symbol in the foremost vertex of $V_g$ that has not yet been added to $G_p$. The oracle begins with a transition system configuration, $C$, initialized with the input sequence, $w_1, ..., w_n$.
%$w_\text{next}$ retains the same meaning in relation to $C$ as in Section~\ref{sec:transition-system}.
%and has access to the corresponding lemmas, $l_1, ..., l_n$. $S_p$, $S_s$ are sets of symbols reserved for promotion-based and sequential generation, respectively.

The oracle is also provided with an approximate alignment, $A = \{(w_i, v_j) \;|\; 1 \leq i \leq n, 1 \leq j \leq m\}$, between the input sequence, $w_{i:n}$, to the nodes in the gold graph, $V_g, |V_g| = m$, which is generated with a greedy matching algorithm. The matching algorithm uses a manually-tuned similarity heuristic built on the superficial similarity of English words, POS, and word order to the stems, suffixes, and preorder positions of the corresponding ULF atoms. A complete description of the alignment algorithm is in the Appendix. This alignment is not necessary to maintain correctness of the oracle, but it is used to cut the losses when the input words become out of sync with the gold graph vertex order.\footnote{When the words become out-of-sync with the gold graph the oracle must rely on \textbf{SymGen} to generate the graph nodes. Since \textbf{SymGen} requires selecting the correct value out the entire vocabulary of ULF atoms, it is much more difficult to predict correctly than \textbf{NameGen}, \textit{TokenGen}, and \textit{LemmaGen} which require only selecting the correct type tag.} Steps 5-7 of the \gen phase uses the alignments to identify whether the buffer or the vertex order is ahead of the other and appropriately sync them back together.

The oracle uses the following procedure, broken down by parsing phase, to extract the action sequence to build the $G_p = G_g$ with $C$ and $A$. %alignment

\vspace{-3mm}
\begin{itemize}[leftmargin=*, noitemsep]
    \item \gen phase: Let $b$ $=$ $\text{Stem}(s_{\text{next}})$, $e$ $=$ $\text{Suffix}(s_{\text{next}})$, $n$ $=$ $\text{IsName}(s_{\text{next}})$.%
    %, and $t_\beta$ $=$ $\text{Next}(\beta)$; $l_\beta$ is the corresponding lemma.
    \footnote{$=$ is string match, $=_i$ is case-insensitive string match, $\text{Pre}$ determines whether its first argument is a prefix of the second and $\text{Pre}_i$ is the case-insensitive counterpart.}
    % \vspace{-2mm}
    \begin{enumerate}[leftmargin=*,noitemsep]
        \item If $n$ and $t_\text{next} = b$, \textbf{NameGen($e$)}
        \item If not $n$ and $t_\text{next} =_i b$, \textbf{TokenGen($e$)}
        \item If not $n$ and $l_\text{next} =_i b$, \textbf{LemmaGen($e$)}
        \item \textbf{MergeBuf} if\\
        $n$ and $\text{Pre}(\text{Concat}(t_\text{next}, \text{`` ''}, t_{\text{next}+1}), b)$ or\\
        not $n$ and $\text{Pre}_i(\text{Concat}(l_\text{next}, \text{``\_''}, l_{\text{next}+1}), b)$ or\\
        not $n$ and $\text{Pre}_i(\text{Concat}(t_\text{next}, \text{``\_''}, t_{\text{next}+1}), b)$
        \item If $(w_i, v_{\text{next}}) \in A$ for $w_i$ before $w_\text{next}$ or $v_{\text{next}} \in S_s$, then \textbf{SymGen($v_{\text{next}}$)} 
        \item If $(w_\text{next}, v_j) \in A$ for $v_j$ which comes after $v_{\text{next}}$ or $v_j \in V_p$, then \textbf{SkipWord}.
        \item Otherwise, \textbf{SymGen($v_{\text{next}}$)}
    \end{enumerate}
    
    Step 5-7 allow the oracle to handle the generation of symbols that are not in word order, by skipping any words that come earlier than the symbol order; and generating symbols that cannot be aligned with \textbf{SymGen} for any reason.
    
    \item \push phase: The push phase of the vanilla cache transition parser's oracle---viz., choosing the cache position whose closest edge into $\beta$ is farthest away---is extended to account not only for direct edges, but also for paths that include only unaligned-symbols.% 
    %Also, give a $-2$ distance score for any cache index where $v_\text{next}$ is a descendent with only unaligned symbols in the path, including $v_\text{next}$ and $\inf$ for any cache index with no aligned descendants in $V_g$, prioritizing the left side of the cache for ties.
    \footnote{
    %PLEASE REWORD ... %% break into separate clauses; I don't know what "these two" are
    The motivation for this is that if only unaligned symbols exist in the path, the full path can be made without changing the relative status of any other node in the transition system. Let $v_1$ and $v_2$ be the end points of the path. With $v_1$ in the cache and the word aligned to $v_2$, $w_{v_2} = w_\text{next}$,  \textbf{SymGen} and \promote can generate all nodes in the path without interacting with the rest of the transition system.    
    %The motivation for this is that if only unaligned symbols exist in the path, a sequence of actions without a change to the status of symbols outside of this path with respect to the transition system these two can be connected if the buffered symbol is $v_\text{next}$ and the cache position is still in the cache. This allows us to effectively treat this path as a single edge in the context of the transition system.
    }
    
    \item \arc phase: The vanilla cache transition system's rule of generating the $ARC$ action for any edge, $e \in E_g \land e \notin E_p$ between the rightmost cache position and the other positions, is extended to also require the child vertex to be fully formed. That is, for the vertex $v_\text{child}$, $|\text{descendants}(v_\text{child}, G_g)| = |\text{descendants}(v_\text{child}, G_p)|$. This enforces bottom-up parsing, which is necessary for both the promotion-based symbol generation and type composition constraint.

    \item \promote phase: If the vertex in the rightmost cache position, $v_{\eta_r}$, is fully formed ($|\text{descendants}(v_{\eta_r}, G_g)| = |\text{descendants}(v_{\eta_r}, G_p)|$) and has a parent node in the \promote lexicon ($\text{label}(\text{parent}(v_{\eta_r}, G_g)) \in S_p$), then the parser generates the action sequence \textbf{PromoteSym}($\text{parent}(v_r, G_g)$), \textbf{PromoteArc}($l_p$) where $l_p$ is the label for the edge from the parent of $v_{\eta_r}$ to $v_{\eta_r}$ in $G_g$ ($\text{EdgeLabel}(\text{parent}(v_{\eta_r}, G_g), v_{\eta_r}, G_g)$).
\vspace{-3mm}
\end{itemize}

\begin{figure*}[t]
\centerline{\includegraphics[width=0.9\linewidth]{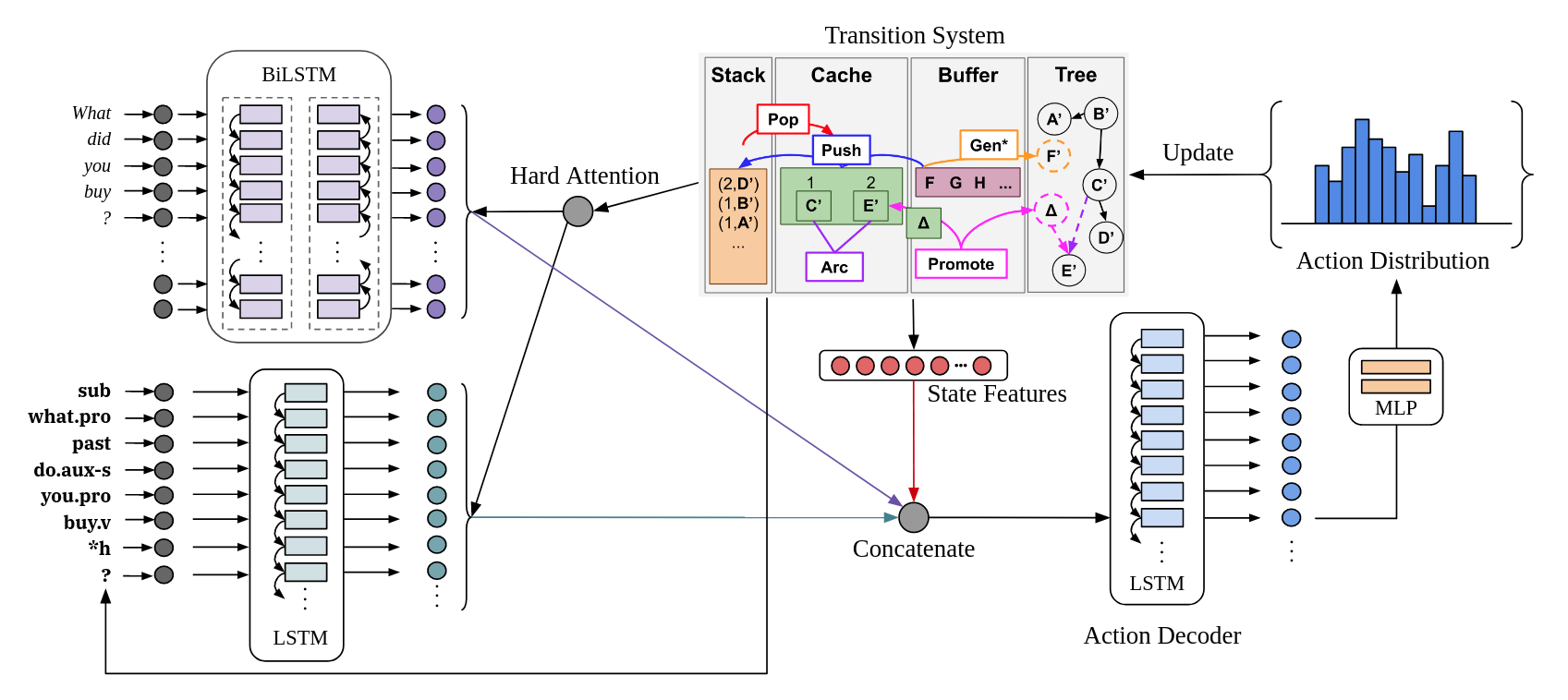}}
\vspace{-3mm}
\caption{The model consists of a sentence-encoding BiLSTM, a symbol-encoding LSTM, and an action-decoding LSTM. New symbols generated in the \gen and \promote phases of the transition system are appended to the symbol sequence. The transition system supplies hard attention pointers that select the relevant word and symbol embeddings. These are concatenated with the transition state feature vector and supplied as input to the action decoder, which predicts the next action that updates the transition system.}
\label{fig:full-model}
\vspace{-4mm}
\end{figure*}

\vspace{-1mm}
\section{Model}
\vspace{-1mm}
% GK: I don't like this naming. We don't really use the full sequence-to-graph model. Rather we just take their embedding layer architectures. I mention in the related works section that the embedding layers are heavily inspired by the stog paper so I think that's enough direct discussion on that front.
%\subsection{Extended sequence-to-graph transduction network}
%Inspired by the sequence-to-graph transduction mechanism in  \citet{zhang-etal-2019-amr}, we extend the AMR node prediction network for ULF symbol prediction.

Our model has three basic components: (1) a word sequence encoder, (2) a ULF atom sequence encoder, and (3) an action decoder, all of which are LSTMs. During decoding, the transition system configuration, $C$, is updated with decoded actions and used to organize the action decoder inputs using the sequence encoders. The system models the following probability
\vspace{-3mm}
\begin{equation}
    \label{eq:model-probability}
    P(a_{1:q}|x_{1:n}) = \prod_{t=1}^q P(a_t|a_{1:t-1},w_{1:n};\theta)
\vspace{-3mm}
\end{equation}
where $a_{1:q}$ is the action sequence, $w_{1:n}$ is the input sequence, and $\theta$ is the set of model parameters. Figure~\ref{fig:full-model} is a diagram of the full model structure.

%\subsection{Sequence Encoders}

%The difference is that each vector is the concatenation of embeddings of GloVe, POS tags and indices, feature vectors from CharCNN.

% GK: we don't do this (I did'nt have time to figure this out and implement it).
%POS tags of nodes are inferred at runtime: if a node is a copy from the input sentence, the POS tag of the corresponding word is used; if a node is a copy from the preceding nodes, the POS tag of its antecedent is used; if a node is a new node emitted from the vocabulary, an $\mathit{UNK}$ tag is used. We exclude BERT embeddings in this layer because ULF symbols, especially their form and order, are significantly different from that of natural language texts (on which BERT was pretrained).
\vspace{-1mm}
\subsection{Word and Symbol Sequence Encoders}
\vspace{-1mm}
The input word embedding sequence $w_{1:n}$ is encoded by a stacked bidirectional LSTM~\cite{hochreiter1997long} with $L_w$ layers. Each word embedding sequence is a concatenation of embeddings of GloVe \cite{pennington-etal-2014-glove}, lemmas, part-of-speech (POS) and named entity (NER) tags, RoBERTa \cite{liu2019roberta}, and features
learned by a character-level convolutional neural
network (CharCNN, \citealp{kim2016character}). As ULF symbols are generated during the parsing process, the symbol embedding sequence $s_{1:m}$, which is the concatenation of a symbol-level learned embedding and the CharCNN feature vector over the symbol string, is encoded by a stacked \textit{uni}directional LSTM of $L_s$ layers.
% \vspace{-3mm}
% \begin{equation}
%     \hb^{L_w}_{w_{1:n}}=\text{StackedBiLSTM}(\hb_{w_{1:n}})
% \end{equation}
% \vspace{-7mm}
% \begin{equation}
%     \hb^{L_s}_{s_{1:m}}=\text{StackedLSTM}(\hb_{s_{1:m}})
% \end{equation}
% \vspace{-10mm}
%\begin{equation}
%    \boldsymbol{h}^{l}_{w_i}=[\overrightarrow{f_w}^{l}(\boldsymbol{h}^{l-1}_{w_i},\boldsymbol{h}^{l}_{w_{i-1}});\overleftarrow{f_w}^{l}(\boldsymbol{h}^{l-1}_{w_i},\boldsymbol{h}^{l}_{w_{i+1}})]
%\end{equation}
%where $\overrightarrow{f_w}^{l}$ and $\overleftarrow{f_w}^{l}$ are $l$-th layer LSTM cells that encode the input sequence in the forward and backward directions respectively; $\boldsymbol{h}^{l}_{w_i}$ is the $l$-th layer encoder hidden state for the $i$-th word, $\boldsymbol{h}^{0}_{w_i}$ is the word embedding layer output for $w_i$.
%
%Since the symbols are generated during the parsing process, the symbol embedding sequence is encoded by a multi-layer \textit{uni}directional LSTM
%\begin{equation}
%    \boldsymbol{h}^{l}_{s_j}=\overrightarrow{f_s}^{l}(\boldsymbol{h}^{l-1}_{s_j},\boldsymbol{h}^{l}_{s_{j-1}})
%\end{equation}
%where $\overrightarrow{f_s}^{l}$ is the $l$-th layer LSTM cell that encodes the symbol sequence. $\boldsymbol{h}^l_{s_j}$ is the $l$-layer encoder hidden state for the $j$-th symbol, and $\boldsymbol{h}^0_{s_j}$ is the symbol embedding layer output for $s_j$.
\vspace{-1mm}
\subsection{Hard Attention}
\vspace{-1mm}
\citet{peng-etal-2018-sequence} found that for AMR parsing with cache transition systems, a hard attention mechanism, tracking the next buffer node position and its aligned word, works better than a soft attention mechanism for selecting the embedding used during decoding. We take this idea and modify the tracking mechanism to find the most relevant word, $w_i$, and symbol, $s_j$, for each phase.
\vspace{-3mm}
\begin{itemize}[leftmargin=*,noitemsep]
    \item \arc and \promote{}*: The symbol $s_j$ in the rightmost cache position and aligned word $w_i$.
    \item \push: The symbol $s_j$ generated in the previous action and aligned word $w_i$.
    \item Otherwise: The last generated symbol $s_j$ and the word $w_i$ in the leftmost $\beta$ position.
\end{itemize}
\vspace{-2mm}
This selects the output sequences $\boldsymbol{h}^{L_w}_{w_i}$ and $\boldsymbol{h}^{L_s}_{s_j}$ from the encoders for the action decoder.
\vspace{-1mm}
\subsection{Transition State Features}
\label{ssec:ts-features}
\vspace{-1mm}

Similar to \citet{peng-etal-2018-sequence}, we extract features from the current transition state configuration, $C$, to feed into the decoder as additional input in the form of learned embeddings
\vspace{-3mm}
\begin{equation}
    \boldsymbol{e}_f(C) = [\boldsymbol{e}_{f_1}(C);\boldsymbol{e}_{f_2}(C);...;\boldsymbol{e}_{f_l}(C)]
\vspace{-3mm}
\end{equation}

\noindent where $\boldsymbol{e}_{f_k}(C) \; (k = 1,...,l)$ is the $k$-th feature embedding, with $l$ total features.
Our features, which are heavily inspired by \citet{peng-etal-2018-sequence}, are as follows.
\vspace{-3mm}
\begin{itemize}[leftmargin=*,noitemsep]
    \item Phase: An indicator of the phase in the transition system.

    \item \pop, \gen features: \textit{Token features}\footnote{The token features are the ULF symbol and the word, lemma, POS, and NER tags of the aligned index of the input.} of the rightmost cache position and the leftmost buffer position; the number of rightward dependency edges from the cache position word and the first three of their labels; and the number of outgoing ULF arcs from the cache position and their first three labels.

    \item \arc, \promote features: For the two cache positions, their token features and the word, symbol\footnote{Symbol distance is based on the order in which the symbols are generated by the parser.}, and dependency distance between them; furthermore, their first three outgoing and single incoming dependency arc labels and their first two outgoing and single incoming ULF arc labels.

    \item \promotearc features: Same as the \promote features but for the rightmost cache position use the node/symbol generated in the preceding \textbf{PromoteSym} action.

    \item \push features: Token features for the leftmost buffer position and all cache positions.
\end{itemize}

%Features that are not relevant in every phase include a NONE value for irrelevant phases.
\vspace{-1mm}
\subsection{Action Encoder/Decoder}
\vspace{-1mm}
The action sequence is encoded by a stacked unidirectional LSTM with $L_a$ layers
% \vspace{-2mm}
% \begin{equation}
%     \hb^{L_a}_{a_{1:q}} = \text{StackedLSTM}(\hb_{a_{1:q}})
% \vspace{-2mm}
% \end{equation}
where the action input embeddings, $\hb_{a_{1:q}}$ are concatenations of the word and symbol encodings.
\vspace{-3mm}
\begin{equation}
    \hb_{a_k} = [\hb_{w_i}^{L_w};\hb_{s_j}^{L_s};\eb_f(C)]
\vspace{-3mm}
\end{equation}
The state features $\boldsymbol{h}^{L_a}_{a_k}$ are then decoded into prediction weights with a linear transformation and ReLU non-linearity.

%is then decoded into unnormalized prediction probabilities, $P_{\Sigma_a}$, of the $(k+1)$-th action
%\vspace{-5mm}
%\begin{equation}
%    P_{\Sigma_a} = \text{ReLU}(\Wb\hb^{L_a}_{a_k} + \bb)
%\vspace{-2mm}
%\end{equation}
%where $\Sigma_a$ is the set of all actions and $\boldsymbol{W}$ and $\boldsymbol{b}$ are learned parameters. 

% The action sequence is encoded by a multi-layer unidirectional LSTM
% \begin{equation}
%     \boldsymbol{h}^{l}_{a_k}=\overrightarrow{f_a}^{l}(\boldsymbol{h}^{l-1}_{a_k},\boldsymbol{h}^{l}_{a_{k-1}})
% \end{equation}
% where $\overrightarrow{f_a}^{l}$ is the $l$-th layer LSTM cell that encodes the action sequence. $\boldsymbol{h}^l_{a_k}$ is the $l$-layer encoder hidden state for the $k$-th action, and $\boldsymbol{h}^0_{a_k}$ is the concatenation of the inputs to the action encoder
% \begin{equation}
%     \boldsymbol{h}^{0}_{a_k}=[\boldsymbol{h}^L_{w_i};\boldsymbol{h}^L_{s_j};\boldsymbol{e}_f(C)].
% \end{equation}
% $\boldsymbol{h}^L_{a_k}$ is then decoded into unnormalized probabilities, $P_{\Sigma_a}$, of the $(k+1)$-th action
% \begin{equation}
%     P_{\Sigma_a} = \text{ReLU}(\boldsymbol{W}\boldsymbol{h}^L_{a_k} + \boldsymbol{b})
% \end{equation}
% where $\Sigma_a$ is the set of all actions and $\boldsymbol{W}$ and $\boldsymbol{b}$ are learned parameters.

\vspace{-1mm}
\section{Parsing}
\vspace{-1mm}
%\subsection{Training and Decoding}
The model is trained on the cross-entropy loss of the model probability \eqref{eq:model-probability} using the oracle action sequence.
%\begin{equation}
%    L = -\sum\limits_{t=1}^q \log P(a_t^*|a_{1:t-1}^*,w_{1:n};\theta)
%\end{equation}
%where $w_{1:n}$ is the input sequence, and $\theta$ is the set of model parameters and extracted features.
Both training and decoding are limited to a maximum action length of 800. For the training set the oracle has an average action length of 134 actions and a maximum action length of 1477.

% TODO(gene): add this table back in if paper is accepted.
%\begin{table*}
%\begin{center}
%{\small
%\begin{tabular}{r|c|l}
%\textbf{ULF Atom} & \textbf{Type} & \textbf{Description} \\
%the.d & $\langle \prd_N, \dom \rangle$ & A function from a unary nominal predicate, $\prd_N$, to an entity, $\dom$. \\
%ka    & $\langle \prd_V, \dom \rangle$ & A function from a unary verbal predicate, $\prd_V$ to an entity. \\
%former.mod-n & $\langle\prd_N,\prd_N\rangle$ & A noun modifier.\\
%mod-n & $\langle \prd, \langle \prd_N, \prd_N \rangle \rangle$ & A function from a unary predicate to a noun modifier.\\
%% (a function from a nominal predicate to another nominal predicate). \\
%\end{tabular}
%}
%\end{center}
%
%\vspace*{-5pt}
%
%\caption{\label{tab:ulf-types}Example ULF types. $\mathcal{N}$ is a shorthand type for unary predicates for brevity, full: $\langle\dom,\langle\sit,\tru\rangle\rangle$.}
%\end{table*}

\subsection{Constrained Decoding}

We investigate two methods of constraining the decoding process with prior knowledge of ULF to overcome the challenge of using a small dataset.
\vspace{-1mm}
\paragraph{ULF Lexicon} To improve symbol generation, we introduce a lexicon with possible ULF atoms for each word. Nouns, verbs, adjectives, adverbs, and preposition entries are automatically converted from the Alvey lexicon~\citep{carroll-grover-1989-derivation} with some manual editing. Pronouns, determiners, and conjunctions entries are extracted from Wiktionary%\footnote{\url{https://en.wiktionary.org/}} 
category lists. Auxiliary verbs entries are manually built from our ULF annotation guidelines. When generating a word-aligned symbol the stem is searched in the lexicon. If the string is present in the lexicon, only corresponding symbols in the lexicon are allowed to be generated.
%The strict variant of the lexicon constraint does not allow the use of \lemmagen and \tokengen at all if stem is not in the lexicon, respectively.
\vspace{-1mm}
\paragraph{Type Composition}
The type system constraint adds a list of types, $T_v$, to accompany $|V_p|$ (the vertices of the partial graph), which stores the ULF type of each vertex. When a vertex, $v$,
is added to $G_p$, its ULF type, $t_v$ is added to $T_v$.
%Table~\ref{tab:ulf-types} provides a few examples of these types. 
This ULF type system is generalized with placeholders for macros and each stage in processing them. 
%An example of the macro types is for the post-nominal modifier macro n+preds, we introduce N+PREDS and +PREDS types. N+PREDS can combine with a nominal predicate to generate +PREDS and +PREDS can either combine with a nominal predicate to remain as +PREDS or be treated as a unary nominal predicate in a composition.
When the parser predicts an arc action during decoding, the types source, $t_s$, and target, $t_t$ nodes are run through a type composition function. If the types can compose, $t_c = (t_s . t_t), t_c \neq \emptyset$, the type of the source node is replaced with $t_c$. Otherwise, the resulting $C$ is not added to the search beam.

\vspace{-1mm}
\section{Experiments}
\vspace{-1mm}
We ran our experiments on a hand-annotated dataset of ULFs totaling 1,738 sentences
(1,378 train, 180 dev, 180 test). The dataset is a mixture of sentences from
crowd-sourced translations, news text, a question dataset, and novels. The
distribution of sentences leans towards more questions, requests, clause-taking
verbs, and counterfactuals because a portion of the dataset comes from the
dataset used by~\citet{kim-etal-2019-generating} for generating inferences from
ULFs of those constructions.
%The data split is done by segmenting the dataset
%into 10 sentence segments and distributing them in a round-robin fashion, with
%the training set receiving eight segments in each round.
The data is split in a chunked round-robin fashion to allow document-level topics to distribute into splits 
while limiting performance inflation due to localized word-choice and grammatical patterns.

\citet{kim-schubert-2019-type} found that interannotator agreement (IA) on ULFs using the \elsmatch metric \citep{kim-schubert-2016-high} is 0.79.\footnote{cf. AMR is reported to have about 0.8 IA using the \smatch metric \citep{tisalos2015slides}} We add a second pass to further reduce variability in our annotations. Further details about the dataset are available in Appendix~\ref{app:dataset} and the complete annotation guidelines are available as part of the dataset.

% TODO: Add dataset description with basic sentence statistics and annotator agreement
\vspace{-1mm}
\paragraph{ULF-AMR}
In order to use parsing and evaluation methods developed for AMR parsing~\cite{banarescu2013abstract}, we rewrite ULFs in penman format~\cite{kasper-1989-flexible} by introducing a node for each ULF atom and generating left-to-right arcs in the order that they appear (:ARG0, :ARG1, etc.), assuming the leftmost constituent is the parent. In order to handle non-atomic operators in penman format which only allows atomic nodes, we introduce a COMPLEX node with an :INSTANCE edge to mark the identity of the non-atomic operator.
\vspace{-1mm}
\paragraph{Setup}
We evaluate the model with \sembleu~\cite{song-gildea-2019-sembleu}, a metric for parsing accuracy of AMRs~\cite{banarescu-etal-2013-abstract}. This metric extends BLEU~\cite{papineni-etal-2002-bleu}
to node- and edge-labeled graphs. We also measure \elsmatch, a generalization of \smatch to graphs with non-atomic nodes, for analysis of the model since it has F1, precision, and recall components.
%We only report results here that are relevant for our discussions. The full tables of results are available in Appendix~\ref{app:full-tables}.

The tokens, lemmas, POS tags, NER tags, and dependencies are all extracted using the Stanford CoreNLP toolkit~\cite{manning-etal-2014-stanford}. In all experiments the model was trained for 25 epochs. Starting at the 12th epoch we measured the \sembleu performance on the dev split with beam size 3.
%Our model is trained with the ADAM optimizer~\cite{kingma2014adam}.
Hyperparameters were tuned manually on the dev split performance of a smaller, preliminary version of the annotation corpus. We use RoBERTa-Base embeddings with frozen parameters, 300 dimensional GloVe embeddings, and 100 dimensional $t_i$, $l_i$, $p_i$, action, and symbol embeddings. The word encoder is 3 layers. The symbol encoder and action decoder are 2 layers.
%All hidden states are 256 dimensions except the symbol encoder which is 128.
Experiments were run on a single NVIDIA Tesla K80 or GeForce RTX 2070 GPU. Training the full model takes about 6 hours. The full tables of results and default parameters are available in Appendix~\ref{app:full-tables}.

\subsection{Results}

\begin{figure}[t]
    \centering
    \includegraphics[width=\linewidth]{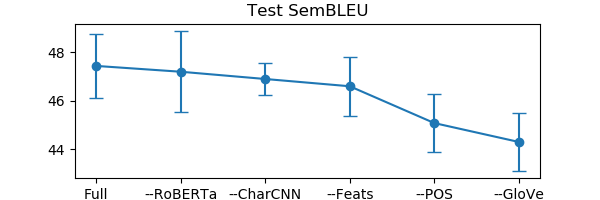}
    \vspace{-7mm}
    \caption{Ablation tests with standard deviation error bars of 5 runs of different random seeds.}
    \label{fig:ablations}
\vspace{-3mm}
\end{figure}
\paragraph{Ablations}

In our ablation tests, the model from the training epoch with the highest dev set \sembleu score is evaluated on the test split with beam size 3.\footnote{Our initial experiments
%took the five models that performed best with a beam size of 3 and
re-evaluated the top-5 choices with a beam size of 10, but we found that the performance consistently degraded and abandoned this step.} The results are shown in Figure~\ref{fig:ablations}.
% GK: not necessary to specify this.
% These ablations omit the constrained decoding methods.

CharCNN and RoBERTa are the least important components---to the point that we cannot conclude that they are of any benefit to the model due to the large overlap in the performance of models with and without them. The GloVe, POS, and feature embeddings are more important. The importance of POS is not surprising given the tight correspondence between POS tags and ULF type tags.
\begin{table}[th]
\begin{center}
\begin{tabular}{l c c}
\hline
\textbf{Model} & {\small\sembleu} & {\small\elsmatch} \\ \hline
\cite{zhang-etal-2019-amr} & 12.3 & 34.3 \\
\cite{cai-lam-2020-amr} & 34.2 & 52.6 \\ \hline
Our best model & \textbf{47.4} & \textbf{59.8} \\
\hline
\end{tabular}
\end{center}
\vspace{-4mm}
\caption{\label{tab:amr-comparison}Comparison to AMR parsers.}
\vspace{-5mm}
\end{table}

\paragraph{Comparison to Baselines}

We compare our parser performance against two AMR parsers with minimal AMR-specific assumptions. The major recent efforts by the research community in AMR parsing make these parsers strong baselines. Specifically, we compare against the sequence-to-graph (STOG) parser~\cite{zhang-etal-2019-amr} and \citeauthor{cai-lam-2020-amr}'s~(\citeyear{cai-lam-2020-amr}) graph-sequence iterative inference~(GS) parser.\footnote{We do not compare our model against the existing rule-based ULF parsers since they are domain specific and cannot handle the range of sentences that appear in our dataset.
} The ULF dataset is preprocessed for these parsers by stripping pipes from names to support the use of a copy mechanism and splitting node labels with spaces into multiple nodes to make the labels compatible with their data pipelines. Table~\ref{tab:amr-comparison} shows the results. The STOG parser fares poorly on both metrics. A review of the results revealed that the parser struggles with node prediction in particular. This is likely the result of the dataset size not properly supporting the parser's latent alignment mechanism.\footnote{The STOG parser is improved by \citep{zhang-etal-2019-broad} with about 1 point of improvement on \smatch. Unfortunately, the code for this parser is not released to the public.} The GS parser performs better than the STOG parser by a large margin, but is still far from our parser's performance. The GS parser also struggles with node prediction, but is more successful in maintaining the correct edges in spite of incorrect node labels.

Investigating the dev set results reveals that our parser is quite successful in node generation, since by design the node generation process reflects the design of ULF atoms. Despite the theoretical capacity to generate node labels without a corresponding uttered word or phrase,
%elided words and logical operators without corresponding words,
our parser only does this for common logical operators such as reifiers and modifier constructors. The GS parser on the other hand, is relatively successful on node labels without uttered correspondences, correctly generating the elided ``you'' in imperatives and the logical operators \ulf{!} and \ulf{multi-sent} which indicate imperatives and multi-sentence annotations, respectively. Our parser also manages to correctly generate a variety of verb phrase constructions, but fails to recognize reified infinitives as arguments of less frequent clausal verbs such as ``neglect'', ``attach'', etc. (as opposed to ``have'', ``tell'') and instead interprets ``to'' as either an argument-marking prepositions or reification of an already reified verb. Examples of parses and a discussion of specific errors are omitted here due to space constraints and provided in Appendix~\ref{app:parse-examples}.
%After running this experiment we became aware of an improvement to the STOG parser, which shows about 1 point of improvement on \smatch\cite{zhang-etal-2019-broad}. This improvement comes from a linking of the node and edge prediction stages and is unlikely to overcome the major performance gap between the STOG parser and our parser on ULF parsing.}

\paragraph{Constrained Decoding}

When evaluating decoding constraints, we select the model by re-running the five best performing epochs with constraints. %models with the constraints on the dev set for the five best performing epochs without constraints. The best performing model out of these is then evaluated on the test split.
When using the type composition constraint, we additionally increase the beam size to 10 so that the parser has backup options when its top choices are filtered out. Table~\ref{tab:constraint-results} presents these results. We see a increase in precision for +Lex, but a greater loss in recall. +Type reduces performance on all metrics.
Due to the bottom-up parsing procedure, a filtering of choices can cascade into fragmented parses. The outputs for an arbitrarily selected run of the model has on average 2.9 fragments per sentence when decoding with the type constraint and 1.4 without. This and the relative performance on the precision metric suggest that constraints improve individual parsing choices, but are too strict, leading to fragmented parses.

\begin{table}[t]
\begin{center}    
\begin{tabular}{l | c | c c c}
\hline
  & \sembleu & \multicolumn{3}{c}{\elsmatch} \\
                &          & {\small F1} & {\small Precision} & {\small Recall} \\ \hline
Full    & \textbf{47.4} & \textbf{59.8} & 60.7 & \textbf{59.0} \\
+Lex    & 46.2 & 57.5 & \textbf{61.5} & 54.1 \\
+Type   & 40.0 & 55.8 & 59.1 & 52.8 \\
\hline
\end{tabular}
\vspace{-2mm}
\caption{\label{tab:constraint-results}Statistics of model performances with constraints added---the average of 5 runs.}
\end{center}
\vspace{-5mm}
\end{table}

\paragraph{Dependence on Length} To investigate the performance dependence on the problem size, we partition the test set into quartiles by oracle action length. The 0 seed of our full model has \sembleu scores of 52, 47, 48, and 31 on the quartiles of increasing length. As expected, the parser performs better on shorter tasks. The parser performance is relatively stable until the last quartile. This is likely due to a long-tail of sentence lengths in our dataset. This last quartile includes sentences with oracle action length ranging from 148 to 1474.

\vspace{-1mm}
\section{Conclusion}
\vspace{-1mm}
We presented the first annotated ULF dataset and the first parser trained on such a dataset.
%We presented the first
%published
%system that learns to parse ULFs from an annotated corpus along with the first annotate ULF dataset.
We showed that our parser is a strong baseline, out-performing existing semantic parsers from a similar task. Surprisingly, our experiments showed that even in this low-resource setting, constrained decoding with a lexicon or a type system does more harm than good. However, the symbol generation method and features designed for ULFs result in a performance lead over using an AMR parser with minimal representational assumptions.

We hope that releasing this dataset will spur other efforts into improving ULF parsing. 
% The size of the dataset will pose a challenge, 
This of course includes expanding the dataset, using our comprehensive annotation guidelines and tools; %suggested by LKS
but we see many
additional % suggested by LKS
avenues of improvement.
% GK: add the following details if the reviewers really want this discussion.
%The type grammar can be used to sample type consistent ULFs which can be converted to English sentences with high-precision~\citep{kim-etal-2019-generating}. This can generate artificial datasets of arbitrary size which can be used to augment the hand-annotated dataset. Language models such as RoBERTa could be used to filter or modify these artificial datasets to avoid low-probability sentences. Another source of improvement could be from better-designed parsing constraints. We presented experiments with strict parsing constraints that lead to incomplete parses.
The type grammar opens up many promising possibilities: sampling of silver data (in conjunction with ULF to English generation~\citep{kim-etal-2019-generating}), use as a weighted constraint, or direct incorporation into a model to avoid the pitfalls we observed in our simple approach to semantic type enforcement.
%Of course, we also hope that there are low-resource NLP methods that we may be unaware of which will help in this task.
% TODO: also advancements of modeling?
% TODO: also better training regimen?

\iffalse
\section{Acknowledgments}
\vspace{-1mm}
This work was supported by NSF EAGER grant NSF IIS-1908595, DARPA CwC subcontract W911NF-15-1-0542, and a Sproull Graduate Fellowship from the University of Rochester. We are grateful to the anonymous reviewers for their helpful feedback.
\fi

\bibliographystyle{acl_natbib}
\bibliography{anthology,acl2021}

\begin{thebibliography}{40}
\expandafter\ifx\csname natexlab\endcsname\relax\def\natexlab#1{#1}\fi

\bibitem[{Abney and Johnson(1991)}]{abney1991memory}
Steven~P Abney and Mark Johnson. 1991.
\newblock Memory requirements and local ambiguities of parsing strategies.
\newblock \emph{Journal of Psycholinguistic Research}, 20(3):233--250.

\bibitem[{Banarescu et~al.(2013{\natexlab{a}})Banarescu, Bonial, Cai,
  Georgescu, Griffitt, Hermjakob, Knight, Koehn, Palmer, and
  Schneider}]{banarescu2013abstract}
Laura Banarescu, Claire Bonial, Shu Cai, Madalina Georgescu, Kira Griffitt, Ulf
  Hermjakob, Kevin Knight, Philipp Koehn, Martha Palmer, and Nathan Schneider.
  2013{\natexlab{a}}.
\newblock Abstract meaning representation for sembanking.
\newblock In \emph{Proceedings of the 7th linguistic annotation workshop and
  interoperability with discourse}, pages 178--186.

\bibitem[{Banarescu et~al.(2013{\natexlab{b}})Banarescu, Bonial, Cai,
  Georgescu, Griffitt, Hermjakob, Knight, Koehn, Palmer, and
  Schneider}]{banarescu-etal-2013-abstract}
Laura Banarescu, Claire Bonial, Shu Cai, Madalina Georgescu, Kira Griffitt, Ulf
  Hermjakob, Kevin Knight, Philipp Koehn, Martha Palmer, and Nathan Schneider.
  2013{\natexlab{b}}.
\newblock \href {https://www.aclweb.org/anthology/W13-2322} {{A}bstract
  {M}eaning {R}epresentation for sembanking}.
\newblock In \emph{Proceedings of the 7th Linguistic Annotation Workshop and
  Interoperability with Discourse}, pages 178--186, Sofia, Bulgaria.
  Association for Computational Linguistics.

\bibitem[{Buys and Blunsom(2017)}]{buys-blunsom-2017-robust}
Jan Buys and Phil Blunsom. 2017.
\newblock \href {https://doi.org/10.18653/v1/P17-1112} {Robust incremental
  neural semantic graph parsing}.
\newblock In \emph{Proceedings of the 55th Annual Meeting of the Association
  for Computational Linguistics (Volume 1: Long Papers)}, pages 1215--1226,
  Vancouver, Canada. Association for Computational Linguistics.

\bibitem[{Cai and Lam(2020)}]{cai-lam-2020-amr}
Deng Cai and Wai Lam. 2020.
\newblock \href {https://doi.org/10.18653/v1/2020.acl-main.119} {{AMR} parsing
  via graph-sequence iterative inference}.
\newblock In \emph{Proceedings of the 58th Annual Meeting of the Association
  for Computational Linguistics}, pages 1290--1301, Online. Association for
  Computational Linguistics.

\bibitem[{Cai and Knight(2013)}]{cai-knight-2013-smatch}
Shu Cai and Kevin Knight. 2013.
\newblock \href {https://www.aclweb.org/anthology/P13-2131} {{S}match: an
  evaluation metric for semantic feature structures}.
\newblock In \emph{Proceedings of the 51st Annual Meeting of the Association
  for Computational Linguistics (Volume 2: Short Papers)}, pages 748--752,
  Sofia, Bulgaria. Association for Computational Linguistics.

\bibitem[{Carroll and Grover(1989)}]{carroll-grover-1989-derivation}
J.~Carroll and C.~Grover. 1989.
\newblock The derivation of a large computational lexicon of english from
  {LDOCE}.
\newblock In Boguraev B. and Briscoe E., editors, \emph{Computational
  Lexicography for Natural Language Processing}, pages 117--134. Longman,
  Harlow, UK.

\bibitem[{Copestake et~al.(2005)Copestake, Flickinger, Pollard, and
  Sag}]{copestake2005RLC}
Ann Copestake, Dan Flickinger, Carl Pollard, and Ivan~A. Sag. 2005.
\newblock {M}inimal {R}ecursion {S}emantics: An introduction.
\newblock \emph{Research on Language and Computation}, 3(2):281--332.

\bibitem[{Damonte et~al.(2017)Damonte, Cohen, and
  Satta}]{damonte-etal-2017-incremental}
Marco Damonte, Shay~B. Cohen, and Giorgio Satta. 2017.
\newblock \href {https://www.aclweb.org/anthology/E17-1051} {An incremental
  parser for {A}bstract {M}eaning {R}epresentation}.
\newblock In \emph{Proceedings of the 15th Conference of the {E}uropean Chapter
  of the Association for Computational Linguistics: Volume 1, Long Papers},
  pages 536--546, Valencia, Spain. Association for Computational Linguistics.

\bibitem[{Gildea et~al.(2018)Gildea, Satta, and Peng}]{gildea-etal-2018-cache}
Daniel Gildea, Giorgio Satta, and Xiaochang Peng. 2018.
\newblock \href {https://doi.org/10.1162/COLI_a_00308} {Cache transition
  systems for graph parsing}.
\newblock \emph{Computational Linguistics}, 44(1):85--118.

\bibitem[{Hershcovich et~al.(2017)Hershcovich, Abend, and
  Rappoport}]{hershcovich-etal-2017-transition}
Daniel Hershcovich, Omri Abend, and Ari Rappoport. 2017.
\newblock \href {https://doi.org/10.18653/v1/P17-1104} {A transition-based
  directed acyclic graph parser for {UCCA}}.
\newblock In \emph{Proceedings of the 55th Annual Meeting of the Association
  for Computational Linguistics (Volume 1: Long Papers)}, pages 1127--1138,
  Vancouver, Canada. Association for Computational Linguistics.

\bibitem[{Hochreiter and Schmidhuber(1997)}]{hochreiter1997long}
Sepp Hochreiter and J{\"u}rgen Schmidhuber. 1997.
\newblock Long short-term memory.
\newblock \emph{Neural computation}, 9(8):1735--1780.

\bibitem[{Kasper(1989)}]{kasper-1989-flexible}
Robert~T. Kasper. 1989.
\newblock \href {https://www.aclweb.org/anthology/H89-1022} {A flexible
  interface for linking applications to {P}enman{'}s sentence generator}.
\newblock In \emph{Speech and Natural Language: Proceedings of a Workshop Held
  at Philadelphia, {P}ennsylvania, {F}ebruary 21-23, 1989}.

\bibitem[{Kim et~al.(2019)Kim, Kane, Duong, Mendiratta, McGuire, Sackstein,
  Platonov, and Schubert}]{kim-etal-2019-generating}
Gene Kim, Benjamin Kane, Viet Duong, Muskaan Mendiratta, Graeme McGuire, Sophie
  Sackstein, Georgiy Platonov, and Lenhart Schubert. 2019.
\newblock \href {https://doi.org/10.18653/v1/W19-3306} {Generating discourse
  inferences from unscoped episodic logical formulas}.
\newblock In \emph{Proceedings of the First International Workshop on Designing
  Meaning Representations}, pages 56--65, Florence, Italy. Association for
  Computational Linguistics.

\bibitem[{Kim and Schubert(2016)}]{kim-schubert-2016-high}
Gene Kim and Lenhart Schubert. 2016.
\newblock \href {https://doi.org/10.18653/v1/S16-2004} {High-fidelity lexical
  axiom construction from verb glosses}.
\newblock In \emph{Proceedings of the Fifth Joint Conference on Lexical and
  Computational Semantics}, pages 34--44, Berlin, Germany. Association for
  Computational Linguistics.

\bibitem[{Kim et~al.(2020)Kim, Juvekar, and Schubert}]{kim-etal-2020-monotonic}
Gene~Louis Kim, Mandar Juvekar, and Lenhart Schubert. 2020.
\newblock Monotonic inference for underspecified episodic logic.
\newblock In \emph{Proceedings of the 1st Workshop on Natural Logic Meets
  Machine Learning (NALOMA)}. Association for Computational Linguistics.

\bibitem[{Kim and Schubert(2019)}]{kim-schubert-2019-type}
Gene~Louis Kim and Lenhart Schubert. 2019.
\newblock \href {https://doi.org/10.18653/v1/W19-0402} {A type-coherent,
  expressive representation as an initial step to language understanding}.
\newblock In \emph{Proceedings of the 13th International Conference on
  Computational Semantics - Long Papers}, pages 13--30, Gothenburg, Sweden.
  Association for Computational Linguistics.

\bibitem[{Kim et~al.(2016)Kim, Jernite, Sontag, and Rush}]{kim2016character}
Yoon Kim, Yacine Jernite, David Sontag, and Alexander~M Rush. 2016.
\newblock Character-aware neural language models.
\newblock In \emph{Thirtieth AAAI Conference on Artificial Intelligence}.

\bibitem[{Lawley et~al.(2019)Lawley, Kim, and
  Schubert}]{lawley-etal-2019-towards}
Lane Lawley, Gene~Louis Kim, and Lenhart Schubert. 2019.
\newblock \href {https://doi.org/10.18653/v1/W19-1102} {Towards natural
  language story understanding with rich logical schemas}.
\newblock In \emph{Proceedings of the Sixth Workshop on Natural Language and
  Computer Science}, pages 11--22, Gothenburg, Sweden. Association for
  Computational Linguistics.

\bibitem[{Li and Roth(2002)}]{li-roth-2002-learning}
Xin Li and Dan Roth. 2002.
\newblock \href {https://www.aclweb.org/anthology/C02-1150} {Learning question
  classifiers}.
\newblock In \emph{{COLING} 2002: The 19th International Conference on
  Computational Linguistics}.

\bibitem[{Liu et~al.(2019)Liu, Ott, Goyal, Du, Joshi, Chen, Levy, Lewis,
  Zettlemoyer, and Stoyanov}]{liu2019roberta}
Yinhan Liu, Myle Ott, Naman Goyal, Jingfei Du, Mandar Joshi, Danqi Chen, Omer
  Levy, Mike Lewis, Luke Zettlemoyer, and Veselin Stoyanov. 2019.
\newblock Ro{BERT}a: A robustly optimized bert pretraining approach.
\newblock \emph{arXiv preprint arXiv:1907.11692}.

\bibitem[{Manning et~al.(2014)Manning, Surdeanu, Bauer, Finkel, Bethard, and
  McClosky}]{manning-etal-2014-stanford}
Christopher Manning, Mihai Surdeanu, John Bauer, Jenny Finkel, Steven Bethard,
  and David McClosky. 2014.
\newblock \href {https://doi.org/10.3115/v1/P14-5010} {The {S}tanford
  {C}ore{NLP} natural language processing toolkit}.
\newblock In \emph{Proceedings of 52nd Annual Meeting of the Association for
  Computational Linguistics: System Demonstrations}, pages 55--60, Baltimore,
  Maryland. Association for Computational Linguistics.

\bibitem[{Montague(1970)}]{montague1970}
Richard Montague. 1970.
\newblock Universal grammar.
\newblock \emph{Theoria}, 36(3):373--398.

\bibitem[{Morbini and Schubert(2009)}]{morbini2009LFCR}
Fabrizio Morbini and Lenhart Schubert. 2009.
\newblock Evaluation of {E}pilog: {A} reasoner for {E}pisodic {L}ogic.
\newblock In \emph{Proceedings of the Ninth International Symposium on Logical
  Formalizations of Commonsense Reasoning}, Toronto, Canada.

\bibitem[{Nivre(2004)}]{nivre2004incrementality}
Joakim Nivre. 2004.
\newblock Incrementality in deterministic dependency parsing.
\newblock In \emph{Proceedings of the workshop on incremental parsing: Bringing
  engineering and cognition together}, pages 50--57.

\bibitem[{Oepen et~al.(2019)Oepen, Abend, Hajic, Hershcovich, Kuhlmann,
  O{'}Gorman, Xue, Chun, Straka, and Uresova}]{oepen-etal-2019-mrp}
Stephan Oepen, Omri Abend, Jan Hajic, Daniel Hershcovich, Marco Kuhlmann, Tim
  O{'}Gorman, Nianwen Xue, Jayeol Chun, Milan Straka, and Zdenka Uresova. 2019.
\newblock \href {https://doi.org/10.18653/v1/K19-2001} {{MRP} 2019:
  Cross-framework meaning representation parsing}.
\newblock In \emph{Proceedings of the Shared Task on Cross-Framework Meaning
  Representation Parsing at the 2019 Conference on Natural Language Learning},
  pages 1--27, Hong Kong. Association for Computational Linguistics.

\bibitem[{Papineni et~al.(2002)Papineni, Roukos, Ward, and
  Zhu}]{papineni-etal-2002-bleu}
Kishore Papineni, Salim Roukos, Todd Ward, and Wei-Jing Zhu. 2002.
\newblock \href {https://doi.org/10.3115/1073083.1073135} {{B}leu: a method for
  automatic evaluation of machine translation}.
\newblock In \emph{Proceedings of the 40th Annual Meeting of the Association
  for Computational Linguistics}, pages 311--318, Philadelphia, Pennsylvania,
  USA. Association for Computational Linguistics.

\bibitem[{Peng et~al.(2018)Peng, Song, Gildea, and
  Satta}]{peng-etal-2018-sequence}
Xiaochang Peng, Linfeng Song, Daniel Gildea, and Giorgio Satta. 2018.
\newblock \href {https://doi.org/10.18653/v1/P18-1171} {Sequence-to-sequence
  models for cache transition systems}.
\newblock In \emph{Proceedings of the 56th Annual Meeting of the Association
  for Computational Linguistics (Volume 1: Long Papers)}, pages 1842--1852,
  Melbourne, Australia. Association for Computational Linguistics.

\bibitem[{Pennington et~al.(2014)Pennington, Socher, and
  Manning}]{pennington-etal-2014-glove}
Jeffrey Pennington, Richard Socher, and Christopher Manning. 2014.
\newblock \href {https://doi.org/10.3115/v1/D14-1162} {{G}lo{V}e: Global
  vectors for word representation}.
\newblock In \emph{Proceedings of the 2014 Conference on Empirical Methods in
  Natural Language Processing ({EMNLP})}, pages 1532--1543, Doha, Qatar.
  Association for Computational Linguistics.

\bibitem[{Platonov et~al.(2020)Platonov, Schubert, Kane, and
  Gindi}]{platonov-etal-2020-spoken}
Georgiy Platonov, Lenhart Schubert, Benjamin Kane, and Aaron Gindi. 2020.
\newblock \href {https://www.aclweb.org/anthology/2020.sigdial-1.16} {A spoken
  dialogue system for spatial question answering in a physical blocks world}.
\newblock In \emph{Proceedings of the 21th Annual Meeting of the Special
  Interest Group on Discourse and Dialogue}, pages 128--131, 1st virtual
  meeting. Association for Computational Linguistics.

\bibitem[{Schubert(2014)}]{schubert2014SP}
Lenhart Schubert. 2014.
\newblock From treebank parses to {E}pisodic {L}ogic and commonsense inference.
\newblock In \emph{Proceedings of the ACL 2014 Workshop on Semantic Parsing},
  pages 55--60, Baltimore, MD. Association for Computational Linguistics.

\bibitem[{Schubert(2015)}]{schubert2015AAAI}
Lenhart Schubert. 2015.
\newblock \href {http://dl.acm.org/citation.cfm?id=2888116.2888296} {Semantic
  representation}.
\newblock In \emph{Proceedings of the Twenty-Ninth AAAI Conference on
  Artificial Intelligence}, AAAI'15, pages 4132--4138. AAAI Press.

\bibitem[{Schubert(2000)}]{schubert2000book2}
Lenhart~K. Schubert. 2000.
\newblock The situations we talk about.
\newblock In Jack Minker, editor, \emph{Logic-based Artificial Intelligence},
  pages 407--439. Kluwer Academic Publishers, Norwell, MA, USA.

\bibitem[{Schubert and Hwang(2000)}]{schubert2000book}
Lenhart~K. Schubert and Chung~Hee Hwang. 2000.
\newblock {E}pisodic {L}ogic meets {L}ittle {R}ed {R}iding {H}ood: A
  comprehensive natural representation for language understanding.
\newblock In Lucja~M. Iwa\'{n}ska and Stuart~C. Shapiro, editors, \emph{Natural
  Language Processing and Knowledge Representation}, pages 111--174. MIT Press,
  Cambridge, MA, USA.

\bibitem[{Song and Gildea(2019)}]{song-gildea-2019-sembleu}
Linfeng Song and Daniel Gildea. 2019.
\newblock \href {https://doi.org/10.18653/v1/P19-1446} {{S}em{B}leu: A robust
  metric for {AMR} parsing evaluation}.
\newblock In \emph{Proceedings of the 57th Annual Meeting of the Association
  for Computational Linguistics}, pages 4547--4552, Florence, Italy.
  Association for Computational Linguistics.

\bibitem[{Tsialos(2015)}]{tisalos2015slides}
Aristeidis Tsialos. 2015.
\newblock \href {www.inf.ed.ac.uk/teaching/courses/tnlp/2014/Aristeidis.pdf}
  {Abstract meaning representation for sembanking}.
\newblock Available at
  \url{www.inf.ed.ac.uk/teaching/courses/tnlp/2014/Aristeidis.pdf}, accessed
  December 8, 2018.

\bibitem[{Wolf(2005)}]{wolf2005thesis}
Florian Wolf. 2005.
\newblock \href {https://dspace.mit.edu/handle/1721.1/28854} {\emph{Coherence
  in natural language : data structures and applications}}.
\newblock Ph.D. thesis, Massachusetts Institute of Technology, Dept. of Brain
  and Cognitive Sciences.

\bibitem[{Zhang et~al.(2016)Zhang, Zhang, and
  Fu}]{zhang-etal-2016-transition-based}
Meishan Zhang, Yue Zhang, and Guohong Fu. 2016.
\newblock \href {https://doi.org/10.18653/v1/P16-1040} {Transition-based neural
  word segmentation}.
\newblock In \emph{Proceedings of the 54th Annual Meeting of the Association
  for Computational Linguistics (Volume 1: Long Papers)}, pages 421--431,
  Berlin, Germany. Association for Computational Linguistics.

\bibitem[{Zhang et~al.(2019{\natexlab{a}})Zhang, Ma, Duh, and
  Van~Durme}]{zhang-etal-2019-amr}
Sheng Zhang, Xutai Ma, Kevin Duh, and Benjamin Van~Durme. 2019{\natexlab{a}}.
\newblock \href {https://doi.org/10.18653/v1/P19-1009} {{AMR} parsing as
  sequence-to-graph transduction}.
\newblock In \emph{Proceedings of the 57th Annual Meeting of the Association
  for Computational Linguistics}, pages 80--94, Florence, Italy. Association
  for Computational Linguistics.

\bibitem[{Zhang et~al.(2019{\natexlab{b}})Zhang, Ma, Duh, and
  Van~Durme}]{zhang-etal-2019-broad}
Sheng Zhang, Xutai Ma, Kevin Duh, and Benjamin Van~Durme. 2019{\natexlab{b}}.
\newblock \href {https://doi.org/10.18653/v1/D19-1392} {Broad-coverage semantic
  parsing as transduction}.
\newblock In \emph{Proceedings of the 2019 Conference on Empirical Methods in
  Natural Language Processing and the 9th International Joint Conference on
  Natural Language Processing (EMNLP-IJCNLP)}, pages 3786--3798, Hong Kong,
  China. Association for Computational Linguistics.

\end{thebibliography}

% The appendix needs to be a separate document. So clear out the page.
\clearpage
\newpage

\appendix
\newpage 

\section{Dataset Details}
\label{app:dataset}

We chose a variety of text sources for constructing this dataset to reduce genre-effects and provide good coverage of all the phenomena we are investigating. Some of these datasets include annotations, which we use only to identify sentence and token boundaries. The dataset includes 1,738 sentences, with a mean, median, min, and max sentence lengths of 10.275, 8, 2, and 128 words, respectively.
%The breakdown of the annotation sourc es are shown in Table~\ref{tab:dataset-breakdown}.

%\begin{table}[h]
%\centering
%\begin{tabular}{ |l|l|l|l|l|l| } 
% \hline
%  \textit{Source} & Count & $\bar{x}$ & $\tilde{x}$ & min & max \\ \hline 
% Tatoeba        & 16xx & & & &\\ 
% DG             & 1xx  & & & &\\
% UIUC QC        & 2xx  & & & &\\
% PG             & 3xx  & & & &\\ \hline
% Total          & 17xx & & & &\\
% \hline
%\end{tabular}
%\caption{\label{tab:dataset-breakdown} Dataset statistics by data source. $\bar{x}$, $\tilde{x}$, min, and max are the mean, median, minimum, and maximum sentence lengths, respectively. DG, PG, and UIUC QC are the Discourse Graphbank, Project Gutenberg, and UIUC Question Classification datasets, respectively.}
%\end{table}

%\begin{figure}[h]
%    \centering
%    \begin{tikzpicture}
%        \pie {62/Tatoeba,
%        26/DG,
%        6/UIUC QC,
%        12/PG}
%    \end{tikzpicture}
%    \caption{Pie chart visualization of the dataset makeup.[TODO: update numbers after final dataset construction.]}
%    \label{fig:dataset-piechart}
%\end{figure}

\subsection{Data Sources}

\begin{itemize}[leftmargin=*]
	\item \textbf{Tatoeba}
    
    The Tatoeba dataset\footnote{\url{https://tatoeba.org/eng/}} consists of crowd-sourced translations from a community-based educational platform. People can request the translation of a sentence from one language to another on the website and other members will provide the translation. Due to this pedagogical structure, the sentences are fluent, simple, and highly-varied. The English portion downloaded on May 18, 2017 contains 687,274 sentences.
    
    \item \textbf{Discourse Graphbank}
    
    The Discourse Graphbank~\cite{wolf2005thesis} is a discourse annotation corpus created from 135 newswire and WSJ texts. We use the discourse annotations to perform sentence delimiting. This dataset is on the order of several thousand sentences.
    
    \item \textbf{Project Gutenberg}
    
    Project Gutenberg\footnote{\url{https://www.gutenberg.org}} is an online repository of texts with expired copyright. We downloaded the top 100 most popular books from the 30 days prior to February 26, 2018. We then ignored books that have non-standard writing styles: poems, plays, archaic texts, instructional books, textbooks, and dictionaries. This collection totals to 578,650 sentences.
    
    \item \textbf{UIUC Question Classification}

	The UIUC Question Classification dataset~\cite{li-roth-2002-learning} consists of questions from the TREC question answering competition. It covers a wide range of question structures on a wide variety of topics, but focuses on factoid questions. This dataset consists of 15,452 questions.
\end{itemize}
Most of the dataset is annotated by random selection of a single or some contiguous sequence of sentences by annotators. However, part of the annotated dataset comes from inference experiments run by~\citet{kim-etal-2019-generating} regarding questions, requests, counterfactuals, and clause-taking verbs. Therefore, the dataset has a bias towards having these phenomena at a higher frequency than expected from a random selection of English text.

A key issue regarding the dataset is its difficulty. We primarily quantify this with the AMR parser baseline, the sequence-to-graph (STOG) parser~\citep{zhang-etal-2019-amr}, in the main text, which performs quite poorly on this dataset. Its performance indicates that the patterns in this dataset are too varied for a modern parsing model to learn without built in ULF-specific biases. Although, part of this is due to the size of the dataset, if the dataset consisted only of short and highly-similar sentences, we would expect a modern neural model, such as the AMR baseline, to be able to learn successful parsing strategy for it.

This reflects the design of the dataset construction. Although the dataset indeed includes many short sentences, especially from the Tatoeba and UIUC Question Classification datasets, the sentences cover a wide range of styles and topics. The Tatoeba dataset is built from a crowd-sourced translation community, so the sentences are not limited in genre and style and has a bias toward sentences that give people trouble when learning a second language. We consider this to be valuable for a parsing dataset since, while the sentences from Tatoeba are usually short, they vary widely in topic and tend to focus on tricky phenomena that give language-learners---and likely parsers---trouble. Sentences from the Discourse Graphbank (news text) and Project Gutenberg (novels) further widen the scope of genres and styles in the dataset. This should make it difficult for a parsing model to overfit to dataset distribution. The dataset also has a considerable representation of longer sentences ($\sim$10\% of the dataset is $>$20 words) including dozens of sentences exceeding 40 words, reaching up to 128 words.

\subsection{Annotation Interface \& Interannotator Agreement}
We use the same annotation interface as \citet{kim-schubert-2019-type}, which includes (1) syntax and bracket highlighting, (2) a sanity checker based on the underlying type grammar, and (3) uncertainty marking to trigger a review by a second annotator. The complete English-to-ULF annotation guideline is attached as a supplementary document.

\citet{kim-schubert-2019-type} reports interannotator agreement (IA) of ULF annotations using this annotation procedure. In summary, they found that agreement among sentences that are marked as \textit{certain} are 0.79 on average and can be up-to 0.88 when we filter for well-trained annotators. For comparison, AMR annotations have been reported to have annotator vs consensus IA of 0.83 for newswire text and 0.79 for webtext using the \textit{smatch} metric~\citep{tisalos2015slides}.

In order to mitigate the issue of low agreement of some annotators in the IA study, each annotation in our dataset was reviewed by a second annotator and corrected if necessary. There was an open discussion among annotators to clear up uncertainty and handle tricky cases during both the original annotation and the reviewing process so the actual dataset annotations are more consistent than the test of IA agreement (which had completely independent annotations) would suggest.

\subsection{Dataset Splits}

The data split is done by segmenting the dataset
into 10 sentence segments and distributing them in a round-robin fashion, with
the training set receiving eight chunks in each round. This splitting method is
designed to allow document-level topics to distribute into splits 
while limiting any performance inflation of the dev and test results that can result 
when localized word-choice and grammatical patterns are distributed into all
splits. 

The Tatoeba dataset further exacerbates the issue of localized word-choice and grammatical
patterns since multiple sentences using the same phrase or grammatical construction often 
appear back-to-back. We suspect that this is because the Tatoeba dataset is ordered chronologically
and users often submit multiple similar sentences in order to help understand a particular
phrase or grammatical pattern in a language that they are learning.

\section{Full ULF Alignment Details}
\label{app:alignment}

The ULF-English alignment system takes into account the similarity of the English word to the ULF atom without the type extension, the similarity of the type extension with the POS tag, and the relative distance of the word and symbol in question.  

Given a sentence $s=w_{1:n}$, which is tokenized, $t_{1:n}$, lemmatized, $l_{1:n}$, and POS tagged, $p_{1:n}$, a set of symbols that are never aligned $S_u$, and a list of ULF atoms $a_{1:m}$, which can be broken up into the base stems, $b_{1:m}$, and suffix extensions, $e_{1:m}$, in order of appearance in the formula (i.e. DFS preorder traversal), the word/atom similarity is defined using the following formulas.

\vspace*{-1em}

{\small\begin{align*}
    &\text{Sim}(w,a) = \max(\text{Olap}(t,b), \text{Olap}(l,b)) \\
    & + 0.5 * (\text{Olap}(p,e) + (1 - |\text{RL}(w,n) - \text{RL}(a,m)|))
\end{align*}}
where token overlap, $\text{Olap}$, is defined as
{\small\begin{equation*}
    \text{Olap}(x,y) = \frac{2*|\text{MaxSharedSubstr}(x,y)|}{|x| + |y|}
\end{equation*}}
and relative location $\text{RL}$ is defined as
{\small\begin{equation*}
    \text{RL}(x,n) = \frac{\text{IndexOf}(x)}{n}
\end{equation*}}
Next, in order of $\text{Sim}(w,a)$, we consider each word-atom pair, $(w_i,a_i)$, $1 \leq i \leq n$ until $\text{Sim}(w,a) < \text{MinSim}$, where $\text{MinSim}$ is set to $1.0$, based on cursory results. We further disregard any alignments that include an atom which shouldn't be aligned~($a_i$ s.t. $a_i \in S_u$). We assume that spans of words align to connected subgraphs, so we cannot accept all word-atom pairs. An word-atom pair, $(w_i, a_i)$, is accepted into the set of token alignments, $A_t$, if and only if the following conditions are met:
\begin{enumerate}
    \item $w_i$ has no alignments or $a_i$ is connected to an atom, $a'$, that is already aligned to $w_i$.
    \item $a_i$ is not in any other alignment or $w_i$ is adjacent to another, $w'$ which is already aligned to $a_i$.
\end{enumerate}

%For a  first we check that either $w_i$ has no alignments or that $a_i$ is connected to one of the atoms $a'$ that is already aligned to $w_i$, second we check that either $a_i$ has no alignments already or that $w_i$ is adjacent to one of the words, $w'$ that is already aligned to $a_i$.  If these to conditions are met, we accept this alignment and add it to the set of alignments.  Otherwise, the pair is discarded.

The token-level (word-atom) alignment, $A_t$, is then converted to connected (span-subgraph) alignment, $A$. This is done with the following algorithm:
\begin{enumerate}
    \item For every atom $a_i$ in one of the aligned pairs of $A_t$, merge all of the words aligned to $a_i$ into a single span, $s_i$. During the initial alignment, we ensured that these words would form a span.
    \item Merge all overlapping spans into single spans and collect the set of atoms that are aligned to each of these spans into a subgraph.\footnote{This can be done in $\mathcal{O}(n \log n)$ time by sorting the spans, then doing a single pass of merging overlapping elements.} These collected subgraphs will be connected because we ensured that for any word the nodes that it is aligned to forms a connected subgraph.
\end{enumerate}

\section{RoBERTa Handling Details}
Except for RoBERTa, all other embeddings are fetched from their corresponding learned embedding lookup tables. RoBERTa uses OpenAI GPT-2 tokenizer for the input sequence and segments words into subwords prior to generating embeddings, which means one input word may correspond to multiple hidden states of RoBERTa.  In order to accurately use these hidden states to represent each word, we apply an average pooling function to the outputs of RoBERTa according to the alignments between the original and GPT-2 tokenized sequences. 

\section{Full Tables}
\label{app:full-tables}

Tables of the full set of raw results and parameters are presented in this section. Table~\ref{tab:ablation-tests} shows the ablations on the model without decoding constraints. This is the basis of Figure~\ref{fig:ablations} in the main text. Table~\ref{tab:lexicon-constraint} shows the performance change with the lexicon constraint and Table~\ref{tab:type-constraint} shows the performance change with the composition constraint. These tables are the basis of Table~\ref{tab:constraint-results} in the main text. Our experiments with the lexicon constraint were more extensive since the type constraint takes considerably longer to run due to requiring a larger beam size and more computational overhead. Table~\ref{tab:model-parameters} presents all of the model parameters in our experiments.

\begin{table*}[!htbp]
\begin{center}
{\small
\begin{tabular}{l||cc||cc|cc|cc}
\hline
\textbf{Ablation} & \multicolumn{2}{c||}{\sembleu} & \multicolumn{6}{c}{\elsmatch} \\ \hline
 & & & \multicolumn{2}{c}{F1} & \multicolumn{2}{c}{Precision} & \multicolumn{2}{c}{Recall}\\
 & Dev & \multicolumn{1}{c||}{Test} & Dev & \multicolumn{1}{c|}{Test} & Dev & \multicolumn{1}{c|}{Test} & Dev & Test \\ \hline
Full     & $46.4\pm1.4$ & $\mathbf{47.4}\pm1.3$ % sembleu 
         & $58.4\pm0.7$ & $\mathbf{59.8}\pm1.0$ % f1
         & $59.1\pm1.1$ & $\mathbf{60.7}\pm1.5$ % precision
         & $57.8\pm0.5$ & $\mathbf{59.0}\pm0.7$ % recall
         \\ \hline
-RoBERTa & $45.5\pm2.4$ & $47.2\pm1.7$ % sembleu 
         & $58.3\pm1.4$ & $59.3\pm1.0$ % f1
         & $59.1\pm1.6$ & $60.5\pm1.1$ % precision
         & $57.5\pm1.2$ & $58.3\pm0.9$ % recall
         \\
-CharCNN & $46.4\pm1.0$ & $46.9\pm0.7$ % sembleu 
         & $58.8\pm0.8$ & $59.3\pm0.4$ % f1
         & $59.4\pm1.3$ & $60.1\pm0.5$ % precision
         & $58.1\pm0.6$ & $58.5\pm0.5$ % recall
         \\
-$e_f(C)$ Feats   & $47.0\pm1.2$ & $46.6\pm1.2$ % sembleu 
         & $58.6\pm0.5$ & $58.8\pm1.1$ % f1
         & $60.4\pm1.2$ & $60.2\pm1.1$ % precision
         & $56.9\pm0.4$ & $57.4\pm1.2$ % recall
         \\
-POS     & $43.8\pm1.7$ & $45.1\pm1.2$ % sembleu 
         & $56.9\pm1.1$ & $58.3\pm1.1$ % f1
         & $56.8\pm1.0$ & $58.7\pm1.1$ % precision
         & $56.9\pm1.2$ & $57.9\pm1.2$ % recall
         \\
-GloVe   & $43.2\pm1.8$ & $44.3\pm1.2$ % sembleu 
         & $56.6\pm1.0$ & $57.1\pm0.9$ % f1
         & $56.9\pm2.7$ & $58.3\pm2.2$ % precision
         & $56.4\pm1.7$ & $56.1\pm2.2$ % recall
         \\
\hline
\end{tabular}
}
\end{center}
\caption{\label{tab:ablation-tests}Ablation results without decoding constraints, mean and standard deviation of 5 runs.
}
\end{table*}

\begin{table*}[!htbp]
\begin{center}
{\small
\begin{tabular}{l||cc||cc|cc|cc}
\hline
\textbf{Ablation} & \multicolumn{2}{c||}{\sembleu} & \multicolumn{6}{c}{\elsmatch} \\ \hline
 & & & \multicolumn{2}{c}{F1} & \multicolumn{2}{c}{Precision} & \multicolumn{2}{c}{Recall}\\
 & Dev & \multicolumn{1}{c||}{Test} & Dev & \multicolumn{1}{c|}{Test} & Dev & \multicolumn{1}{c|}{Test} & Dev & Test \\ \hline
Full     & $47.3\pm0.6$ & $46.2\pm0.3$ % sembleu 
         & $56.3\pm0.7$ & $57.5\pm0.8$ % f1
         & $60.2\pm0.5$ & $61.5\pm1.2$ % precision
         & $52.9\pm0.9$ & $54.1\pm1.5$ % recall
         \\ 
$\Delta\bar{x}$
         & & -1.2 & & -2.3 & & +0.8 & & -4.9 \\
         \hline
-RoBERTa & $46.6\pm1.3$ & $46.9\pm0.6$ % sembleu 
         & $56.1\pm0.6$ & $57.8\pm0.4$ % f1
         & $60.0\pm0.7$ & $60.5\pm0.9$ % precision
         & $52.6\pm0.6$ & $55.3\pm0.5$ % recall
         \\
-CharCNN & $45.8\pm2.3$ & $45.5\pm2.5$ % sembleu 
         & $56.1\pm1.4$ & $56.9\pm1.1$ % f1
         & $59.3\pm2.4$ & $59.6\pm1.8$ % precision
         & $53.3\pm1.1$ & $54.5\pm1.5$ % recall
         \\
-$e_f(C)$ Feats   
         & $45.9\pm1.5$ & $45.6\pm0.9$ % sembleu 
         & $56.5\pm0.6$ & $57.0\pm0.5$ % f1
         & $62.0\pm0.8$ & $61.4\pm0.6$ % precision
         & $52.0\pm1.1$ & $53.3\pm0.5$ % recall
         \\
-POS     & $44.1\pm2.0$ & $44.5\pm0.9$ % sembleu 
         & $55.3\pm0.2$ & $56.6\pm0.7$ % f1
         & $58.5\pm2.2$ & $60.4\pm0.8$ % precision
         & $52.6\pm2.3$ & $53.2\pm1.4$ % recall
         \\
-GloVe   & $46.1\pm1.1$ & $45.4\pm1.4$ % sembleu 
         & $55.9\pm0.9$ & $57.0\pm0.6$ % f1
         & $59.5\pm1.5$ & $60.3\pm0.8$ % precision
         & $52.7\pm1.0$ & $54.0\pm0.7$ % recall
         \\
\hline
\end{tabular}
}
\end{center}
\caption{\label{tab:lexicon-constraint}Ablation results with the lexicon constraint, mean and standard deviation of 5 runs. $\Delta\bar{x}$ is the difference in the mean score between the test set results of the model with the lexicon constraint and without, i.e. Table~\ref{tab:ablation-tests}. We only list this for the full model, but the pattern of higher precision but lower scores on other metrics generally holds for the other variants as well.}
\end{table*}

\begin{table*}[!htbp]
\begin{center}
{\small
\begin{tabular}{l||cc||cc|cc|cc}
\hline
\textbf{Ablation} & \multicolumn{2}{c||}{\sembleu} & \multicolumn{6}{c}{\elsmatch} \\ \hline
 & & & \multicolumn{2}{c}{F1} & \multicolumn{2}{c}{Precision} & \multicolumn{2}{c}{Recall}\\
 & Dev & \multicolumn{1}{c||}{Test} & Dev & \multicolumn{1}{c|}{Test} & Dev & \multicolumn{1}{c|}{Test} & Dev & Test \\ \hline
Full     & $38.3\pm2.3$ & $40.0\pm1.4$ % sembleu 
         & $54.2\pm1.2$ & $55.8\pm1.2$ % f1
         & $57.6\pm1.0$ & $59.1\pm1.2$ % precision
         & $51.1\pm1.5$ & $52.8\pm1.4$ % recall
         \\ 
$\Delta\bar{x}$
         & & -7.4 & & -4.0 & & -1.6 & & -6.2 \\
\hline
\end{tabular}
}
\end{center}
\caption{\label{tab:type-constraint}Ablation results with the type composition constraint, mean and standard deviation of 5 runs. $\Delta\bar{x}$ is the difference in the mean score between the test set results of the model with the type constraint and without, i.e. Table~\ref{tab:ablation-tests}. We only ran the full model for this test because this constraint takes much longer to run.}
\end{table*}

\begin{table*}[htbp]
\begin{center}
\begin{tabular}{l c c}
\hline
\textbf{Model} & \multicolumn{2}{c}{Fragments/Sentence}
\\
 & $\alpha$ & $\tau$ \\ \hline
Full      & 1.4 & 2.9 \\
-CharCNN  & 1.1 & 3.5 \\
-$e_f(C)$ Feats & 1.4 & 3.9 \\
-POS & 1.5 & 3.2 \\ \hline
$\bar{x}$ & 1.4 & 3.4 \\
\hline
\end{tabular}
\end{center}
\caption{Fragments per sentence on the test set decoding results for a subset of the ablated lexicon-constrained models (Table~\ref{tab:lexicon-constraint}). $\alpha$ is the original model and $\tau$ is with the type composition constraint.}
\end{table*}

\begin{table*}[ht]
\begin{center}
\begin{tabular}{l r}
\hline\hline
\textbf{GloVe.840B.300d embeddings} \\
dim & 300 \\ 
\hline\hline
\textbf{RoBERTa embeddings} \\
source & RoBERTa-Base \\ 
dim & 768 \\ 
\hline\hline
\textbf{POS tag embeddings} \\
dim & 100 \\ 
\hline\hline
\textbf{Lemma embeddings} \\
dim & 100 \\ 
\hline\hline
\textbf{CharCNN} \\
num\_filters & 100 \\ 
ngram\_filter\_sizes & [3] \\
\hline\hline
\textbf{Action embeddings} \\
dim & 100 \\ 
\hline\hline
\textbf{Transition system feature embeddings} \\
dim & 25 \\ 
\hline\hline
\textbf{Word encoder} \\
hidden\_size & 256 \\
num\_layers & 3 \\
\hline\hline
\textbf{Symbol encoder} \\
hidden\_size & 128 \\
num\_layers & 2 \\
\hline\hline
\textbf{Action decoder} \\
hidden\_size & 256 \\
num\_layers & 2 \\
\hline\hline
\textbf{MLP decoder} \\
hidden\_size & 256 \\
activation\_function & ReLU\\
num\_layers & 1 \\
\hline\hline
\textbf{Optimizer} \\
type & ADAM \\
learning\_rate & 0.001 \\
max\_grad\_norm & 5.0 \\
dropout & 0.33 \\
num\_epochs & 25 \\
\hline\hline
\textbf{Beam size} \\
without type composition filtering & 3 \\
with type composition filtering & 10 \\
\hline\hline
\textbf{Vocabulary}\\
word\_encoder\_vocab\_ size & 9200\\
symbol\_encoder\_vocab\_ size & 7300 \\
\hline\hline
\textbf{Batch size} & 32 \\
\end{tabular}
\end{center}
\caption{\label{tab:model-parameters}Default model parameters.}
\end{table*}

\section{Parse Examples}
\label{app:parse-examples}

Figure~\ref{fig:parse-examples} shows six parse examples of our parser and the GS parser in reference to the gold standard. Generally, we find that our parser does much better on node generation for nodes that correspond to an input word. For example, the GS parser on example 1 uses \textit{(plur *s)} for the word ``speech'' and \textit{iron.n} for the words ``silver'' and  ``silence''. This isn't to say that our parser doesn't make mistakes. But the mistakes are not as open-ended. For example, our parser mistakenly annotates ``silver'' as a noun in example 1 when in fact it should be an adjective (compared against ``golden''). The GS parser seems to pick the closest word in its vocabulary, which is generated from the training set and closed. This leads to strange annotations like \textit{iron.n} for the word ``silence''. If there is nothing close available, then it can derail the entire parse. In example 4, the GS parser is unable to find a node label for the word ``device'' which derails the parse to generate \textit{(mod-n (mod-n man.n) (mod-n man.n iron.n) mod-n mod-n)} for the text span ``device is attached firmly to the ceiling''.

This isn't to say that the GS parser always performs worse than our parser. When it comes to words that are elided (\textit{\{you\}.pro} in example 4), nodes generated from multiple words (\textit{had\_better.aux-s} in example 3), or logical symbols unassociated with a particular word (\textit{multi-sent} in example 6), the GS parser consistently performs better than our parser. Our parser has no special mechanism for these handling these cases and prefers to avoid generating node labels without an anchoring word.

A common mistake by our parser seems to be nested reifiers, which is not possible in the EL type system (e.g. \textit{(to (ka come.v))} in example 5 and \textit{(to (ka (show.v ..)))} in example 6). Other common mistakes that could be fixed by type coherence enforcement is mistakenly shifting a term into a modifier (e.g. \textit{(adv-a (to ...))} in example 6). In the EL type system only predicates can be shifted into modifiers. 

\sethlcolor{red}

\begin{figure*}[t]

\begin{enumerate}[itemsep=1em]
\item {\large \textit{``Speech is silver but silence is golden.''}}

\textbf{Gold:} \ulf{(((k speech.n) ((pres be.v) silver.a)) but.cc ((k silence.n) ((pres be.v) golden.a)))}

\vspace{0.5em}

\textbf{Ours:} \ulf{(((k speech.n) ((pres be.v) silver\textcolor{red}{.n})) \hl{ } (k silence.n) ((pres be.v) golden.a))}

\vspace{0.5em}

\textbf{GS:} \ulf{(((k \textcolor{red}{(plur *s)}) ((pres be.v) \textcolor{red}{(= (k iron.n))})) but.cc (\textcolor{red}{(k iron.n)} ((pres be.v) \textcolor{red}{=})))}

\item {\large \textit{``You neglected to tell me to buy bread.''}}

\textbf{Gold:} \ulf{(you.pro ((past neglect.v) (to (tell.v me.pro (to (buy.v (k bread.n)))))))}

\vspace{0.5em}

\textbf{Ours:} \ulf{(you.pro ((past neglect.v) \textcolor{red}{(adv-e} (to (tell.v me.pro (to (buy.v (k bread.n)))))))}

\vspace{0.5em}

\textbf{GS:} \ulf{(you.pro ((past \textcolor{red}{fail.v}) (to (tell.v me.pro \textcolor{red}{\{ref\}.pro}))))}

\item {\large \textit{``You'd better knuckle down to work.''}}

\textbf{Gold:} \ulf{(you.pro ((pres had\_better.aux-s) (knuckle.v down.adv-a (to work.v))))}

\vspace{0.5em}

\textbf{Ours:} \ulf{(you.pro \textcolor{red}{((pres would.aux-s) }(knuckle.v \textcolor{red}{down.a (adv-a (to.p} work.v)))))}

\vspace{0.5em}

\textbf{GS:} \ulf{(you.pro ((pres had\_better.aux-s) \textcolor{red}{(go.v (to.p-arg (k work.n)) (adv-a (to.p (ka} work.v))))))}

\item {\large \textit{``Make sure that the device is attached firmly to the ceiling.''}}

\textbf{Gold:} \ulf{(\{you\}.pro ((pres make.v) sure.a\\
\hspace*{3em}(that ((the.d device.n) ((pres (pasv attach.v)) firmly.adv-a (to.p-arg (the.d ceiling.n)))))))}

\vspace{0.5em}

\textbf{Ours:} \ulf{(\hl{ } ((pres make.v) sure.a \textcolor{red}{that.pro (tht}\\ 
\hspace*{3em}((the.d device.n) ((pres \textcolor{red}{be.v}) \hl{ } \textcolor{red}{(k (n+preds} attach.v \textcolor{red}{(to.p-arg} \hl{ } ceiling.n))))))))}

\vspace{0.5em}

\textbf{GS:} \ulf{((\{you\}.pro ((pres make.v) (sure.a\\
\hspace*{3em}(that (the.d \textcolor{red}{(mod-n (mod-n man.n) (mod-n man.n iron.n) mod-n mod-n)}))))) !)}

\item {\large \textit{``Can't I persuade you to come?''}}

\textbf{Gold:} \ulf{(((pres can.aux-v) not i.pro (persuade.v you.pro (to come.v)) ?)}

\vspace{0.5em}

\textbf{Ours:} \ulf{\textcolor{red}{(sub} ((pres can.aux-v) not i.pro (persuade.v you.pro (to \textcolor{red}{(ka} come.v)) ?)\textcolor{red}{))}}

\vspace{0.5em}

\textbf{GS:} \ulf{(((pres can.aux-v) not i.pro (\hl{ } \textcolor{red}{come.v} (to come.v) \textcolor{red}{you.pro})) ?)}

\item {\large \textit{``Look carefully. I'm going to show you how it's done.''}}

\textbf{Gold:} \ulf{(multi-sent ((\{you\}.pro ((pres look.v) carefully.adv-a)) !)\\
\hspace*{3em}(i.pro ((pres be-going-to.aux-v) \\
\hspace*{6em}(show.v you.pro (ans-to (sub how.pq (it.pro ((pres (pasv do.v)) *h))))))))}

\vspace{0.5em}

\textbf{Ours:} \ulf{(\hl{ }((pres look.v) carefully.adv-a) \hl{ }\\
\hspace*{3em}(\textcolor{red}{tht} (i.pro (\textcolor{red}{(pres be.v) (go.v}\\
\hspace*{6em}\textcolor{red}{(adv-a (to (ka} (show.v you.pro (sub how.pq (it.pro ((pres be.v) \textcolor{red}{do.n} \hl{ }))))))))))))}

\vspace{0.5em}

\textbf{GS:} \ulf{(multi-sent ((\{you\}.pro \textcolor{red}{((pres be.v) you.pro fine.a)) !)}\\
\hspace*{3em}(i.pro ((pres \textcolor{red}{be.aux-v) (go.v (to (do.v} you.pro *h))))))}

\end{enumerate}

\caption{Several parse examples comparing behavior of our parser with the stronger baseline, the GS parser. For each example, the top is the gold parse, the center is our parser, and the bottom is the GS~\cite{cai-lam-2020-amr} parser. Errors are written in red. If something from the gold parse is omitted, a red highlighted block marks the location.}
\label{fig:parse-examples}
\end{figure*}

\end{document}